\documentclass[10pt,journal,compsoc]{IEEEtran}
\ifCLASSOPTIONcompsoc
  \usepackage[nocompress]{cite}
\else
  \usepackage{cite}
\fi
\ifCLASSINFOpdf
\else
\fi

\usepackage{microtype}
\usepackage{graphicx}
\usepackage{booktabs} 

\usepackage{hyperref}
\usepackage{url}

\usepackage{balance}

\usepackage{amsmath}
\usepackage{amssymb}
\usepackage{mathtools}
\usepackage{amsthm}

\usepackage[capitalize,noabbrev]{cleveref}

\theoremstyle{plain}

\newtheorem{theorem}{Theorem}[section]
\newtheorem{proposition}[theorem]{Proposition}

\theoremstyle{definition}

\newtheorem{assumption}[theorem]{Assumption}
\theoremstyle{remark}

\usepackage[textsize=tiny]{todonotes}

\usepackage[utf8]{inputenc} 
\usepackage[T1]{fontenc}    
\usepackage{hyperref}       
\usepackage{url}            
\usepackage{booktabs}       
\usepackage{amsfonts}       
\usepackage{nicefrac}       
\usepackage{microtype}      
\usepackage{amsmath}
\usepackage{graphicx}
\usepackage{xcolor}
\usepackage{colortbl}
\usepackage{multirow}
\usepackage{float}
\usepackage{subcaption}

\newcommand{\bi}{\begin{itemize}}
\newcommand{\ei}{\end{itemize}}
\usepackage{enumitem}
\setlist[itemize]{leftmargin=*}
\setlist[enumerate]{leftmargin=*}

\usepackage{tikz}
\def\checkmark{\tikz\fill[scale=0.4](0,.35) -- (.25,0) -- (1,.7) -- (.25,.15) -- cycle;} 

\usepackage{array}
\newcolumntype{P}[1]{>{\centering\arraybackslash}p{#1}}

\usepackage{makecell}
\usepackage[linesnumbered,ruled,vlined]{algorithm2e}
\usepackage{algorithmic}
\usepackage[skins]{tcolorbox}

\newenvironment{RQ}{\vspace{2mm}\begin{tcolorbox}[enhanced,width=\linewidth,size=fbox,colback=blue!5,drop shadow southeast,sharp corners]}{\end{tcolorbox}}

\begin{document}
%
\title{FairBalance: How to Achieve Equalized Odds With Data Pre-processing}
%

%
%
%
%

\author{Zhe Yu,~\IEEEmembership{Member,~IEEE,}
        Joymallya Chakraborty,
        and Tim Menzies,~\IEEEmembership{Fellow,~IEEE}
\IEEEcompsocitemizethanks{\IEEEcompsocthanksitem Zhe Yu is with the Department
of Software Engineering, Rochester Institute of Technology.\protect\\
E-mail: zxyvse@rit.edu
\IEEEcompsocthanksitem Joymallya Chakraborty is with Amazon.
\IEEEcompsocthanksitem Tim Menzies is with the Department of Computer Science, North Carolina State University.}
}

\IEEEtitleabstractindextext{%
\begin{abstract}

This research seeks to benefit the software engineering society by providing a simple yet effective pre-processing approach to achieve equalized odds fairness in machine learning software. 
Fairness issues have attracted increasing attention since machine learning software is increasingly used for high-stakes and high-risk decisions. It is the responsibility of all software developers to make their software accountable by ensuring that the machine learning software do not perform differently on different sensitive demographic groups--- satisfying equalized odds. Different from prior works which either optimize for an equalized odds related metric during the learning process like a black-box, or manipulate the training data following some intuition; this work studies the root cause of the violation of equalized odds and how to tackle it. We found that equalizing the class distribution in each demographic group with sample weights is a necessary condition for achieving equalized odds without modifying the normal training process. In addition, an important partial condition for equalized odds (zero average odds difference) can be guaranteed when the class distributions are weighted to be not only equal but also balanced (1:1). Based on these analyses, we proposed FairBalance, a pre-processing algorithm which balances the class distribution in each demographic group by assigning calculated weights to the training data. On eight real-world datasets, our empirical results show that, at low computational overhead, the proposed pre-processing algorithm FairBalance can significantly improve equalized odds without much, if any damage to the utility. FairBalance also outperforms existing state-of-the-art approaches in terms of equalized odds. To facilitate reuse, reproduction, and validation, we made our scripts available at \url{https://github.com/hil-se/FairBalance}.
\end{abstract}

}

\maketitle

\IEEEdisplaynontitleabstractindextext

%
\IEEEpeerreviewmaketitle

\IEEEraisesectionheading{\section{Introduction}\label{sec:introduction}}

Increasingly, machine learning and artificial intelligence software is being used to make decisions that affect people's lives. This has raised much concern on the fairness of that kind of reasoning. Decision making software can be ``biased'';
i.e. it gives undue advantage to specific group of people (where those groups are determined by sex, race, etc.). Such bias in the machine learning software can have serious consequences in deciding whether a patient gets released from the hospital~\cite{Medical_Diagnosis,7473150}, which loan applications are approved~\cite{forbes}, which citizens get bail or sentenced to jail~\cite{propublica}, who gets admitted/hired by universities/companies~\cite{amazon}.  
With such prevalence of the potentially biased machine learning software being developed, it is the responsibility of all software developers to make their software accountable by reducing the unwanted biases from machine learning software predictions.  

A machine learning software can have different types of biases. (1) It can inherit the bias from its training data labels, e.g. a machine learning software will predict that all female applicants should not be hired if it learns from biased hiring decisions where no female applicants (even when they were qualified) were hired historically. (2) It can also favor one demographic group over another by generating more accurate or positive predictions on data from that group, e.g. it has been found that, in 2020, the face recognition software from large companies including Amazon, Microsoft, IBM, etc. predict in significantly lower accuracy ($20-30\%$) for darker female faces than for lighter male faces~\cite{face}. Another example is the COMPAS analysis~\cite{propublica} where a machine learning software has similar accuracy across black and white groups in predicting whether a defendant will re-offend in two years. However, the software has around 20\% higher false positive rate and around 20\% lower false negative rate in predicting a black defendant will re-offend--- i.e. more black defendants were wrongly predicted as higher risk when they actually won't and more white defendants were wrongly predicted as lower risk when they will re-offend. Given that the definition and criteria of fair decisions vary from context to context~\cite{abu2020contextual}, it is not the software developers' responsibility to decide whether the training data labels are fair (which is the responsibility of the domain experts). On the other hand, with the assumption that the training data labels are correct and fair, software developers should ensure that the machine learning software do not perform differently on different sensitive demographic groups. That is, the true positive rate and false positive rate of the predictions on each demographic group should be the same for a fairly designed machine learning software~\cite{chakraborty2020fairway,chakraborty2021bias,peng2022fairmask}.
Thus, amongst the various fairness notions proposed for different scenarios~\cite{verma2018fairness,mehrabi2021survey}, this work specifically targets equalized odds~\cite{hardt2016equality}. It is a simple, interpretable, and easily checkable notion of nondiscrimination with respect to a specified sensitive attribute~\cite{hardt2016equality}. Most importantly, it always allows for the perfectly accurate solution--- the model's predictions always equal to the ground truth labels. Equalized odds are almost always applied to evaluate fairness in machine learning software when ground truth labels are available.   However, most existing machine learning fairness solutions do not directly target equalized odds, nor do they analyze how and why equalized odds can be achieved. 

In addition, most existing machine learning fairness solutions only affect one sensitive attribute (e.g. sex) at a time. For example, on a dataset with two sensitive attributes sex and race, most existing approaches can learn either a fair model on sex or a fair model on race, but not a fair model on both sex and race~\cite{kamiran2012data,calmon2017optimized,zhang2018mitigating}. This also hinders the application of the fairness algorithms since a fair machine learning model cannot be biased on any sensitive attribute. Some in-processing bias mitigation algorithms can tackle multiple sensitive attributes at the same time by optimizing for both utility and specific fairness metrics~\cite{lowy2021fermi} (including equalized odds). However, such in-processing algorithms are usually very expensive. Magic parameters also need to be decided beforehand to trade off between utility and fairness metrics. In addition, these in-processing methods usually limit the models used for the decision making.

To sum up, prior works either optimize for an equalized odds related metric during the learning process like a black-box~\cite{lowy2021fermi,li2022achieving}, or manipulate the training data following some intuition~\cite{kamiran2012data,chakraborty2021bias}. None of the work studies the root cause of the violation of equalized odds and how to tackle it. To bridge this gap, we analyzed the conditions behind equalized odds and derived two important conditions: (1) a necessary condition of pre-training sample weights to achieve equalized odds, and (2) a sufficient condition of pre-training sample weights to satisfy zero average odds difference (a partial/relaxed condition for equalized odds) in the training data. Such analyses provided the theoretical foundation for our proposed pre-processing algorithm FairBalance. These conditions suggest that, 

\noindent\textit{The violation of equalized odds of the learned model is positively related to the weighted class distribution differences across each demographic group in the training data.} 

 Back to the COMPAS analysis example, in the training data, the ratio of black defendants re-offended is higher than that of white defendants. Such difference in class distribution caused the learned software to have a higher false positive rate and lower false negative rate on black defendants.   Satisfying both conditions, the proposed pre-processing algorithm FairBalance adjusts the sample weights of training data from each demographic group so that the weighted class distributions across each demographic group become balanced. With the empirical results on eight real world datasets, we show that, as a simple yet effective pre-processing algorithm, FairBalance guarantees zero smAOD (smoothed maximum average odds difference defined later in Section~\ref{sec:Methodology}) in the training data, can handle multiple sensitive attributes simultaneously, has low computational overhead ($O(n)$), has little damage to utility, and is model-agnostic.


The overall \textbf{contributions} of this paper include:
\begin{itemize}
\item
We analyzed the conditions of equalized odds and derived two important conditions for achieving equalized odds by adjusting sample weights of the training data.
\item
 We proposed our pre-processing algorithm satisfying the necessary and sufficient conditions to directly target equalized odds of multiple sensitive attributes simultaneously.  
\item
With empirical results on eight datasets, we tested the proposed algorithm. FairBalance significantly outperformed existing state-of-the-art fairness approaches in terms of equalized odds. It also has little damage to utility and low computational overhead ($O(n)$).
\item
 We demonstrated the generalizability of FairBalance by achieving equalized odds with a complex deep neural network VGG-16 on a real world image processing dataset.  
\item
To facilitate reuse, reproduction, and validation of this work, our scripts and data are available at \url{https://github.com/hil-se/FairBalance}.
\end{itemize}

The rest of this paper is structured as follows. Section~\ref{sect:Related Work} provides the background and related work of this paper. Section~\ref{sec:Methodology} analyzes the conditions of equalized odds and proposes our pre-processing algorithms based on the conditions. To test the proposed algorithms, Section~\ref{sect:Experiments} presents the empirical experiment setups on eight datasets while Section~\ref{sec:results} shows the experiment results and answers the research questions. Followed by discussion of threats to validity in Section~\ref{threats} and conclusion in Section~\ref{sect:Conclusion}.

\subsection{Notations}
\label{sec:notation}
Here, we summarize the general notations applied to the rest of the paper. Consider a binary classification problem,
\bi
\item
$A\in\mathbb{R}^q$ represents the sensitive attributes.
\item
$X\in\mathbb{R}^p$ represents the independent variables excluding the sensitive attributes.
\item
$Y\in\{0, 1\}$ represents the binary dependent variable.
\item
$f_{\theta}(X,A)$ is a predictor of trainable parameters $\theta$ which takes inputs of $X$ and $A$.
\item
$\hat{Y}\in\{0, 1\}$ is the binary output of the predictor $f_{\theta}(X,A)$.
\item
$w(A=a,Y=y)$ is the calculated sample weight for training data points with $A=a, Y=y$.
\ei

\section{Background and Related Work}
\label{sect:Related Work}

Ethical bias in machine learning software is a well-known and fast-growing topic. It leads to unfair treatments to people belonging to certain groups.
Recently, large industries have started putting more and more importance on ethical issues of machine learning model and software. 
IEEE \cite{IEEEethics}, the European Union~\cite{EU}, and Microsoft~\cite{MicrosoftEthics} have each recently published principles for ethical AI conduct. All three stated that intelligent systems or machine learning software must be fair when used in real-life applications. IBM launched an extensible open-source software toolkit called AI Fairness 360~\cite{IBM}
to help detect and mitigate bias in machine learning models throughout the application life cycle. Microsoft has created a research group called FATE~\cite{FATE} (Fairness, Accountability, Transparency, and Ethics in AI). Facebook announced they developed a tool called Fairness Flow~\cite{Fairness_Flow} that can determine whether a ML algorithm is biased or not. ASE 2019 has organized first International Workshop on Explainable Software~\cite{EXPLAIN} where issues of ethical AI were extensively discussed. 

Various different fairness notions~\cite{verma2018fairness,mehrabi2021survey} have been defined to assess whether a trained machine learning model has ethical bias. Most of these fairness notions, e.g. individual fairness, fairness through awareness, and demographic parity, test both bias emerged in the learning process and bias inherited from the training labels~\cite{sharma2020data}. In this work, we focus on mitigating the bias emerged in the learning process and assume that all training data and their labels are perfectly correct. Under this assumption, a perfect predictor $\hat{Y}=Y$ should always be fair and unbiased.  Many of the popular fairness testing metrics, e.g. FlipTest~\cite{black2020fliptest,zheng2022neuronfair,zhang2020white}, individual fairness violation~\cite{zhang2021efficient}, and demographic parity~\cite{jiang2022generalized}, do not always allow a perfect predictor to be evaluated as fair and unbiased when the sensitive attributes are indeed correlated to the dependent variable~\cite{friedler2021possibility}. For example, in the COMPAS analysis, the re-offended rate in the male group is indeed larger than that in the female group--- $P(Y=\text{Re-offended}|A=\text{Male})>P(Y=\text{Re-offended}|A=\text{Female})$. A perfect predictor $\hat{Y} = Y$ will have the same predicted re-offended rates--- $P(\hat{Y}=\text{Re-offended}|A=\text{Male})=P(Y=\text{Re-offended}|A=\text{Male})>P(Y=\text{Re-offended}|A=\text{Female})=P(\hat{Y}=\text{Re-offended}|A=\text{Female})$. Therefore, it would not satisfy demographic parity or FlipTest but will achieve equalized odds.
This is why we specifically target equalized odds in the experiments.  

\subsection{Equalized Odds}
\label{sect:equal_odds}

As defined by Hardt et al.~\cite{hardt2016equality}, a predictor $\hat{Y}$ satisfies equalized odds with respect to sensitive attribute $A$ and outcome $Y$, if $\hat{Y}$ and $A$ are conditionally independent on $Y$. More specifically, for binary targets $Y$ and sensitive attributes $A$, equalized odds is equivalent to:
\begin{equation}\label{eqodds}
\begin{aligned}
&P(\hat{Y}=1 | A=0, Y=y)\\
=&P(\hat{Y}=1 | A=1, Y=y),\quad y\in{0,1}
\end{aligned}
\end{equation}
The above equation also means that the predictor has the same true positive rate and false positive rate across the two demographics $A=0$ and $A=1$. Equalized odds thus enforces both equal bias and equal accuracy in all demographics, punishing models that perform well only on the majority. 

Equalized odds is a widely applied fairness notation since it always allows for the perfectly accurate solution of $\hat{Y} = Y$. More broadly, the criterion of equalized odds is easier to achieve the more accurate the predictor $\hat{Y}$ is, aligning fairness with the central goal in supervised learning of building more accurate predictors. It is important to note that there is no single best fairness notion for every scenario, only the most appropriate fairness notion for the scenario under study. Two major limitations of equalized odds are
\bi
\item
\textbf{ It heavily relies on the correctness of training data labels}. Thus it can be misleading when the training data labels themselves are biased and discriminative.
\item
\textbf{It ignores the underlying causal structures of the data that actually generate disparities.} When the underlying causal structures are known, it is more appropriate to use counterfactual fairness notions~\cite{zhang2018equality} where the causal structure is being utilized to ensure that the sensitive attributes are not the causes for the disparities of predictions. Counterfactual fairness would reflect unfairness in the training data labels as well.   
\ei
In the scenarios studied by this work, we assume the correctness of the training data labels--- they correctly reflect the distribution of test data labels--- and that the causal relationships are unknown.  

To measure the extent to which a predictor satisfies equalized odds, two important fairness metrics were established:
\begin{itemize}
\item 
\textbf{Average Odds Difference (AOD)}~\cite{pleiss2017fairness, IBM}: Average of difference in False Positive Rates (FPR) and True Positive Rates (TPR)~\eqref{eq:aod}.
\begin{equation}
\label{eq:aod}
\begin{aligned}
    AOD = &0.5\times[(FPR(A=0) - FPR(A=1))\\
    &+ (TPR(A=0) - TPR(A=1))]
\end{aligned}
\end{equation}
\item
\textbf{Equal Opportunity Difference (EOD)}~\cite{hardt2016equality}:  Difference of True Positive Rates (TPR)~\eqref{eq:eod}. 
\begin{equation}
\label{eq:eod}
    EOD = TPR(A=0) - TPR(A=1)
\end{equation}
\end{itemize}
Where TPR and FPR are the true positive rate and false positive rate calculated as \eqref{eq1}.  
\begin{equation}
\label{eq1}
\begin{aligned}
    TPR(A=a_k) &= P(\hat{Y}=1|A=a_k,Y=1)\\
    FPR(A=a_k) &= P(\hat{Y}=1|A=a_k,Y=0)\\
    \forall a_k &\in \{0,1\}.
\end{aligned}
\end{equation}  


The two fairness metrics AOD and EOD each features a relaxed version of equalized odds. When $AOD=0$, the sums of true positive rate and false positive rate are the same across the two demographics $A=0$ and $A=1$. This metric measures whether the predictor $\hat{Y}$ favors one demographic over the other. When $EOD=0$, the true positive rates are the same across the two demographics $A=0$ and $A=1$. This metric measures a relaxed version of equalized odds called equal opportunity where only true positive rates were considered. When both $AOD=0$ and $EOD=0$, perfect equalized odds will be achieved.

\subsection{Fairness on Multiple Sensitive Attributes}
\label{sec:multiple}

Most machine learning fairness research only considers one sensitive attribute with binary values (such as the definition of equalized odds by Hardt et al.~\cite{hardt2016equality}). However, it is very important to extend the fairness notions to multiple sensitive attributes. This is because Intersectionality is a critical lens for analyzing how unfair processes in society affect certain groups~\cite{foulds2020bayesian}. In many real-world scenarios, multiple sensitive attributes exist and discrimination against any subgroup $a_i=(a^{(1)}_i,a^{(2)}_i,\cdots,a^{(q)}_i)$ is not desired. Here, $a_i\in A$ where $A$ is the set of all possible combinations of the sensitive attributes $a^{(1)},a^{(2)},\cdots,a^{(q)}$. For example, when there are two sensitive attributes $a^{(1)} = \{\text{Male}, \text{Female}\}$ and $a^{(2)}=\{\text{White}, \text{Non-White}\}$, the demographic groups are $A=\{(\text{Male}, \text{White}), (\text{Male}, \text{Non-White}), (\text{Female}, \text{White}),$ (\text{Female}, \text{Non-White})$\}$. Following this notation, equalized odds on multiple sensitive attributes is equivalent to:
\begin{equation}\label{eq:modds}
\begin{aligned}
&P(\hat{Y}=1 | A=a_i, Y=y)\\
=&P(\hat{Y}=1 | A=a_j, Y=y),\, \forall a_i, a_j\in A, y\in{0,1}
\end{aligned}
\end{equation}
Based on \eqref{eq:modds}, the following two metrics shown in \eqref{eq:maod} and \eqref{eq:meod} evaluate the violation of equalized odds on multiple sensitive attributes:
\begin{equation}
\label{eq:maod}
\begin{aligned}
    mAOD = &0.5\times[\max_{a_i\in A}(TPR(A=a_i)+FPR(A=a_i))\\
    &- \min_{a_j\in A}(TPR(A=a_j)+FPR(A=a_j))]
\end{aligned}
\end{equation}
\begin{equation}
\label{eq:meod}
    mEOD = \max_{a_i\in A}TPR(A=a_i) - \min_{a_j\in A}TPR(A=a_j)
\end{equation}

\subsection{Related Work}
\label{sec:related work}

Prior work on machine learning fairness can be classified into three types depending on when the treatments are applied:

\textbf{Pre-processing algorithms.} Training data is pre-processed in such a way that discrimination or bias is reduced before training the model. Overall, there are three main categories of pre-processing algorithms to reduce machine learning bias: 
\bi
\item
\textbf{Category 1} features algorithms modifying the values of training data points (including feature values, sensitive attribute values, and label values). For example, Feldman et al.~\cite{feldman2015certifying} designed \textit{disparate impact remover} which edits feature values to increase group fairness while preserving rank-ordering within groups. Calmon et al.~\cite{NIPS2017_6988} proposed an \textit{optimized pre-processing} method which learns a probabilistic transformation that edits the labels and features with individual distortion and group fairness. Another pre-processing technique, \textit{learning fair representations}, finds a latent representation which encodes the data well but obfuscates information about sensitive attributes~\cite{zemel2013learning}. Romano et al.~\cite{romano2020achieving} replace the original sensitive attributes with values independent from the labels $Y$ to train a model approximately achieving equalized odds. Similarly, Peng et al.~\cite{peng2022fairmask} replace the sensitive attributes with values predicted based on other attributes. 
\item
\textbf{Category 2} algorithms aim to increase training efficacy by removing certain data points from the training data. For example, Chakraborty et al. proposed Fairway~\cite{chakraborty2020fairway} and FairSituation~\cite{chakraborty2021bias} which select a subset of the original data for training by performing different tests on the original training data points. 
\item
\textbf{Category 3} algorithms manipulate training data distribution by either adjusting the sample weights or oversample data points from certain demographics. For example, Kamiran and Calders~\cite{kamiran2012data} proposed \textit{reweighing} method that generates weights for the training examples in each (group, label) combination differently to achieve fairness. Fair-SMOTE~\cite{chakraborty2021bias} oversamples training data points from minority groups with synthetic data points~\cite{chawla2002smote} to achieve balanced class distributions. Similarly, Yan et al.~\cite{yan2020fair} also oversample training data points from minority groups with synthetic data points to achieve balanced class distributions. However, it focused on the scenario where sensitive attributes are unknown and applied a clustering method to identify different demographic groups in an unsupervised manner. For the actual fairness improvement part, Yan et al.~\cite{yan2020fair} is the same as Fair-SMOTE~\cite{chakraborty2021bias} as they both apply the SMOTE~\cite{chawla2002smote} algorithm to oversample minority class data to match the number of the majority class data in every demographic group.
\ei

\textbf{In-processing algorithms.} These approaches adjust the way a machine learning model is trained to reduce the bias. Zhang et al.~\cite{zhang2018mitigating} proposed \textit{Adversarial debiasing} method which learns a classifier to increase accuracy and simultaneously reduce an adversary's ability to determine the sensitive attribute from the predictions. This leads to generation of fair classifier because the predictions cannot carry any group discrimination information that the adversary can exploit. Celis et al.~\cite{celis2019classification} designed a \textit{meta algorithm} to take the fairness metric as part of the input and return a classifier optimized with respect to that fairness metric. Kamishima et al.~\cite{10.1007/978-3-642-33486-3_3}  developed \textit{Prejudice Remover} technique which adds a discrimination-aware regularization term to the learning objective of the classifier. Li and Liu~\cite{li2022achieving} tunes the sample weight for each training data point so that a specific fairness notion such as equal opportunity can be achieved along with the best prediction accuracy on a validation set. Several approaches~\cite{zafar2017fairness,agarwal2018reductions,thomas2019preventing,lowy2021fermi,kearns2018preventing} solve the problem as a constrained optimization problem by adding a constraint of a certain bias metric to the loss function and optimizes it. Among these, Lowy et al.~\cite{lowy2021fermi} measured fairness violation using exponential Rényi mutual information (ERMI) and designed an in-processing algorithm to reduce ERMI and prediction errors with stochastic optimization. There are also works manipulating the way deep neural networks are trained by dropping out certain neurons related to the sensitive attributes~\cite{gao2022fairneuron}.

\begin{table*}[!tb]
\caption{Characteristics of each treatment.}
\label{tab:treatments}
\centering
\setlength\tabcolsep{2pt}
\begin{tabular}{|p{0.15\textwidth}|P{0.16\textwidth} P{0.16\textwidth} P{0.16\textwidth} P{0.16\textwidth} P{0.15\textwidth}|}
\hline
\textbf{Treatment} & Satisfies the necessary condition for equalized odds & Satisfies the sufficient condition for $smAOD=0$ & Keeps size difference across the demographic groups & Removes confusing training data & Synthetic training data
\\ \hline
None &  &  &  & &  \\
Fairway~\cite{chakraborty2020fairway} &  &  &  & \checkmark &  \\
FairSituation~\cite{chakraborty2021bias} &  &  &  & \checkmark &  \\
Fair-SMOTE~\cite{chakraborty2021bias} & \checkmark & \checkmark & & & \checkmark\\
Reweighing~\cite{kamiran2012data} & \checkmark & & \checkmark & &\\
\textbf{FairBalance} & \checkmark & \checkmark & \checkmark & &\\
\textbf{FairBalanceVariant} & \checkmark & \checkmark &  & &\\\hline
\end{tabular}
\end{table*}

\textbf{Post-processing algorithms.} These approaches adjust the prediction threshold after the model is trained to reduce specific fairness metrics. Kamiran et al.~\cite{Kamiran:2018:ERO:3165328.3165686} proposed \textit{Reject option classification} which gives favorable outcomes to unprivileged groups and unfavorable outcomes to privileged groups within a confidence band around the decision boundary with the highest uncertainty. \textit{Equalized odds post-processing}~\cite{pleiss2017fairness,hardt2016equality,awasthi2020equalized} specifically finds the optimal thresholds of an existing predictor to achieve equal opportunity or equalized odds. Such post-processing algorithms usually do not change the prediction probabilities (the ROC curve will stay the same) but only selects different thresholds for the classification. A simple baseline approach Fairea~\cite{hort2021fairea} even randomly mutates the predictions of certain classes to a different class. 

\textbf{Ensemble algorithms.} These approaches combine different bias mitigation methods/models~\cite{chakraborty2021bias,bhaskaruni2019improving,iosifidis2019fae,kenfack2021impact,chen2022maat} to address fairness bugs. For example, Chen et al.~\cite{chen2022maat} train two separate models, one optimized for fairness and one optimized for accuracy, then the average of the two models' outputs are utilized for the final prediction.

In this paper, we focus on the pre-processing approaches since they are usually model-agnostic and cost-effective. Also, based on the analyses later in Section~\ref{sec:Methodology}, the class distributions in each demographic group are the main factor affecting equalized odds and pre-processing is the most efficient and effective way to change that. The in-processing and post-processing algorithms are indirect and costly in terms of equalized odds. In addition, under the assumption that all the data values are correct, we avoided \textbf{Category 1} algorithms since they will modify the data values and possibly mislead the learned models. Algorithms in \textbf{Category 3} is also preferred over those in \textbf{Category 2} since \textbf{Category 2} algorithms do not fully utilize the entire training data. Therefore, later in Section~\ref{sect:Experiments}, we will compare the proposed algorithms FairBalance and FairBalanceVariant with two baseline pre-processing algorithms Fair-SMOTE~\cite{chakraborty2021bias} and Reweighing~\cite{kamiran2012data} in \textbf{Category 3}, two baseline pre-processing algorithms Fairway~\cite{chakraborty2020fairway} and FairSituation~\cite{chakraborty2021bias} in \textbf{Category 2}, and one baseline None without any fairness treatment. A preview of the differences between each treatments studied in this paper can be found in Table~\ref{tab:treatments}. Details of each algorithm will be provided in Section~\ref{fairbalance}, Section~\ref{exist} and Section~\ref{RQs}.

\section{Methodology}
\label{sec:Methodology}

Existing work showed that, adjusting the sample weights of training data points affects the model's fairness the most~\cite{biswas2021fair,kamiran2012data}. Therefore, we aim to achieve equalized odds by adjusting the weight on the training data points. For simplicity, we define our problem under the following two assumptions:

\begin{assumption}\label{Assumption 1}
Labels in the training data and test data follow the same distribution--- a perfect predictor trained on the training data will have 100\% accuracy on the test data.
\end{assumption}

\begin{assumption}\label{Assumption 2}
The data distribution in each demographic group $a\in A$ is independent.
\end{assumption}

\subsection{Problem Statement}
\label{sec:problem}

Given a set of labeled data $(X\in\mathbb{R}^p, A\in\mathbb{R}^q, Y\in\{0, 1\})$ following the Assumption~\ref{Assumption 1} and \ref{Assumption 2}, we aim to learn a predictor shown in \eqref{eq:predictor},
\begin{equation}\label{eq:predictor}
\hat{Y} = \left\{
\begin{matrix}
1 & \text{if }f_{\theta}(X, A)\ge 0.5\\
0 & \text{if }f_{\theta}(X, A)< 0.5
\end{matrix}
\right.
\end{equation}
that satisfies equalized odds defined in \eqref{eq:modds}. The predictor is learned by minimizing the weighted loss function with parameter $\theta$:
\begin{equation}\label{eq:loss}
\begin{aligned}
\mathcal{L}(\theta) = \frac{1}{N}\sum^{N}_{i=1} w(a_i, y_i)\cdot E(f_{\theta}(x_i,a_i), y_i).
\end{aligned}
\end{equation}
Where $w(a_i,y_i)$ is the weight on the $i_{th}$ data point. $E(f_{\theta(X,A)}, y)$ is a specific loss such as binary cross-entropy or squared error. Without loss of generality, we use logistic regression as the predictor so that
\begin{equation}\label{eq:logistic_loss}
\begin{aligned}
&E(f_{\theta(X,A)}, y)\\
= &- [y\cdot\log f_{\theta}(X, A)+(1-y)\cdot\log (1-f_{\theta}(X, A))]
\end{aligned}
\end{equation}
where
\begin{equation}\label{eq:logistic}
\begin{aligned}
f_{\theta}(X, A)= \frac{1}{1+\exp^{-z}}
\end{aligned}
\end{equation}
and
\begin{equation}\label{eq:logistic_z}
\begin{aligned}
z = &\theta^{(0)}+\theta^{(1)}x^{(1)}+\cdots +\theta^{(p)}x^{(p)}\\
&+\theta^{(p+1)}a^{(p+1)}+\cdots +\theta^{(p+q)}a^{(p+q)}.
\end{aligned}
\end{equation}

\subsection{Smoothed Metrics}\label{sec:smooth}

To better understand the relationship between the weight $w(A, Y)$ and equalized odds, we analyze the smoothed version of mAOD and mEOD as shown in \eqref{eq:smaod} and \eqref{eq:smeod}.
\begin{equation}\scriptsize
\label{eq:smaod}
\begin{aligned}
    &smAOD\\
    = &0.5\times[\max_{a_i\in A}(P(\hat{Y}=1 | A=a_i, Y=1)+P(\hat{Y}=1 | A=a_i, Y=0))\\
    &- \min_{a_j\in A}(P(\hat{Y}=1 | A=a_j, Y=1)+P(\hat{Y}=1 | A=a_j, Y=0))]\\
    = &0.5\times[\max_{a_i\in A} (\frac{\sum\limits_{A=a_i, Y=1}f_{\theta}(X, A)}{|A=a_i, Y=1|} + \frac{\sum\limits_{A=a_i, Y=0}f_{\theta}(X, A)}{|A=a_i, Y=0|})\\
    &- \min_{a_j\in A}(\frac{\sum\limits_{A=a_j, Y=1}f_{\theta}(X, A)}{|A=a_j, Y=1|} + \frac{\sum\limits_{A=a_j, Y=0}f_{\theta}(X, A)}{|A=a_j, Y=0|}).
\end{aligned}
\end{equation}
\begin{equation}\scriptsize
\label{eq:smeod}
    smEOD = \max_{a_i\in A}\frac{\sum\limits_{A=a_i, Y=1}f_{\theta}(X, A)}{|A=a_i, Y=1|} - \min_{a_j\in A}\frac{\sum\limits_{A=a_j, Y=1}f_{\theta}(X, A)}{|A=a_j, Y=1|}.
\end{equation}

 Given that the predictor $f_{\theta}(X, A) \in [0, 1]$ is a continuous output of the probability of the predicted data point belongs to Class $Y=1$, it more accurately reflect the prediction of the classification model and in many scenarios, this continuous output is being used as the final decisions instead of the discrete prediction $\hat{Y}$. Thus, the smoothed metrics $smAOD$ and $smEOD$ better evaluate the violation of equalized odds in \eqref{eq:modds}. Meanwhile, to evaluate equalized odds when using the discrete predictions $\hat{Y}$ as the final decisions, we will also show each treatment's performance in terms of $mAOD$ and $mEOD$ in our experiments along with $smAOD$ and $smEOD$ in Section~\ref{sect:Experiments} and \ref{sec:results}.  

\subsection{Necessary Condition}

\begin{proposition}\label{necessary}
The necessary condition for achieving equalized odds ($smAOD=0$ and $smEOD=0$) is 
\begin{equation}\label{eq:necessary}
\forall a_k \in A, \quad \frac{w(A=a_k, Y=1)}{w(A=a_k, Y=0)}=\alpha\frac{|A=a_k, Y=0|}{|A=a_k, Y=1|}
\end{equation}
where $\alpha \in \mathbb{R}_{\ge 0}$ is a positive constant.
\end{proposition}
That is, the weighted class distribution in each demographic group $a_k \in A$ should be the same: $$\frac{w(A=a_k, Y=1)|A=a_k, Y=1|}{w(A=a_k, Y=0)|A=a_k, Y=0|}=\alpha.$$
\begin{proof}
Given $smAOD=0$ and $smEOD=0$, we have
\begin{equation}\label{eq:eo}
\begin{aligned}
&\frac{\sum\limits_{A=a_k, Y=1}f_{\theta}(X, A)}{|A=a_k, Y=1|}= c_1\text{ and} \\
& \frac{\sum\limits_{A=a_k, Y=0}f_{\theta}(X, A)}{|A=a_k, Y=0|}= c_0, \quad \forall a_k \in A.
\end{aligned}
\end{equation}
where $c_1, c_0 \in [0,1]$ are two positive constants. The learned (sub-)optimal model $f_{\theta}(X, A)$ should satisfy
\begin{equation}\label{eq:c1}
\frac{\partial \mathcal{L}(\theta)}{\partial \theta} = 0.
\end{equation}
Apply \eqref{eq:loss}, \eqref{eq:logistic_loss}, \eqref{eq:logistic}, and \eqref{eq:logistic_z} to \eqref{eq:c1} we have 
\begin{equation}\label{eq:c2}
\begin{aligned}
&\frac{\partial \mathcal{L}(\theta)}{\partial \theta^{(0)}} \\
= &\frac{1}{N}\sum^{N}_{i=1} w(a_i,y_i) \frac{\partial E(f_{\theta}(x_i,a_i), y_i)}{\partial \theta^{(0)}}\\
= & \frac{1}{N}\sum^{N}_{i=1} w(a_i,y_i) (f_{\theta}(x_i, a_i)-y_i)\\
= & \frac{1}{N}\sum\limits_{a_k \in A} \left( w(a_k, 1) \cdot \sum\limits_{A=a_k, Y=1} (f_{\theta}(X, A)-1)\right.\\
&+ \left. w(a_k, 0) \cdot \sum\limits_{A=a_k, Y=0} f_{\theta}(X, A) \right)\\
= & 0
\end{aligned}
\end{equation}
Since the data distribution in each demographic group $a_k \in A$ is independent according to Assumption~\ref{Assumption 2}, we have
\begin{equation}\label{eq:c3}
\begin{aligned}
&  w(a_k, 1) \cdot \sum\limits_{A=a_k, Y=1} (f_{\theta}(X, A)-1)&\\
&+ w(a_k, 0) \cdot \sum\limits_{A=a_k, Y=0} f_{\theta}(X, A) &\\
&=  0 , &\forall a_k \in A.
\end{aligned}
\end{equation}
Apply \eqref{eq:eo} to \eqref{eq:c3} we have
\begin{equation}\label{eq:c4}
\begin{aligned}
\forall a_k \in A, \quad & w(a_k, 1) \cdot (c_1-1) \cdot |A=a_k, Y=1|\\
&+w(a_k, 0)\cdot c_0 \cdot |A=a_k, Y=0| =0.
\end{aligned}
\end{equation}
Therefore we have \eqref{eq:necessary} with $\alpha = \frac{c_0}{1-c_1} \in \mathbb{R}_{\ge 0}$.
\end{proof}

Proposition~\ref{necessary} explains why a machine learning model trained with uniform sample weights will always violates equalized odds when the class distributions $\frac{|A=a_k, Y=0|}{|A=a_k, Y=1|}$ are different in each demographics $a_k \in A$.

\textbf{Generalizability: }Note that, although the analysis is performed on a logistic regression classifier as specified in Section~\ref{sec:problem}, \eqref{eq:c2} holds for any unbiased predictor with zero mean of training errors and an intercept term $\theta^{(0)}$.  This property will be demonstrated in RQ5 of Section~\ref{sect:Experiments} and \ref{sec:results}.  

\subsection{Sufficient Condition}

\begin{proposition}\label{sufficient}
One sufficient condition for $smAOD=0$ is $\alpha = 1$ in \eqref{eq:necessary}.
\end{proposition}
That is, the weighted class distribution in each demographic group $a_k \in A$ should be perfectly balanced: $$\frac{w(A=a_k, Y=1)|A=a_k, Y=1|}{w(A=a_k, Y=0)|A=a_k, Y=0|}=1.$$
\begin{proof}
With $\alpha = 1$ in \eqref{eq:necessary} we have
\begin{equation}\label{eq:sufficient}
\forall a_k \in A, \quad \frac{w(A=a_k, Y=1)}{w(A=a_k, Y=0)}=\frac{|A=a_k, Y=0|}{|A=a_k, Y=1|}
\end{equation}
Apply \eqref{eq:sufficient} to \eqref{eq:c3} we have
\begin{equation}\label{eq:c5}
\begin{aligned}
&  w(a_k, 0)|A=a_k, Y=0|\left(\frac{\sum\limits_{A=a_k, Y=1} (f_{\theta}(X, A)-1)}{|A=a_k, Y=1|}\right.\\
&+ \left.\frac{ \sum\limits_{A=a_k, Y=0} f_{\theta}(X, A)}{|A=a_k, Y=0|} \right)=0, \quad \forall a_k \in A.
\end{aligned}
\end{equation}
This guarantees $smAOD=0$ since $\forall a_k \in A$, 
$$\frac{\sum\limits_{A=a_k, Y=1} f_{\theta}(X, A)}{|A=a_k, Y=1|} + \frac{ \sum\limits_{A=a_k, Y=0} f_{\theta}(X, A)}{|A=a_k, Y=0|} = 1.$$
\end{proof}

\subsection{Analyses of Existing Algorithms}\label{exist}

Existing pre-processing treatments in \textbf{Category 3} fit into our problem statement in Section~\ref{sec:problem} and can be analyzed for whether they satisfy the necessary condition and the sufficient condition.

\textbf{Reweighing:} Perfectly falling into the problem statement in Section~\ref{sec:problem}, the Reweighing~\cite{kamiran2012data} algorithm sets the sample weight $w(A, Y)$ as \eqref{eq:reweighing}.
\begin{equation}\label{eq:reweighing}
\begin{aligned}
&w_{RW}(A=a_k, Y=y_i) = \frac{|A=a_k|\cdot|Y=y_i|}{|A=a_k, Y=y_i|}\\
&\forall a_k \in A, \quad \forall y_i \in Y
\end{aligned}
\end{equation}
While Reweighing satisfies the necessary condition in Proposition~\ref{necessary}:
\begin{equation*}
\begin{aligned}
&\frac{w_{RW}(A=a_k, Y=1)}{w_{RW}(A=a_k, Y=0)} = \frac{|Y=1|}{|Y=0|}\cdot \frac{|A=a_k, Y=0|}{|A=a_k, Y=1|}\\
&\forall a_k \in A.
\end{aligned}
\end{equation*}
It does not satisfy the sufficient condition in Proposition~\ref{sufficient} when
\begin{equation*}
\begin{aligned}
\alpha = \frac{|Y=1|}{|Y=0|} \neq 1.
\end{aligned}
\end{equation*}
As a result, it is possible for Reweighing~\cite{kamiran2012data} to achieve equalized odds, but there is no guarantee for it to achieve $smAOD=0$.

\textbf{Fair-SMOTE: }As for Fair-SMOTE~\cite{chakraborty2021bias}, it oversamples the training data to $(X', A', Y')$ so that
\begin{equation}\label{eq:Fair-SMOTE}
\begin{aligned}
&|A'=a_k, Y'=y_i| = |A'=a_l, Y'=y_j|\\
&\forall a_k, a_l \in A', \quad \forall y_i, y_j \in Y'
\end{aligned}
\end{equation}
Then, with unit weights $w(A',Y')=1$, it satisfies both the necessary condition in Proposition~\ref{necessary} and the sufficient condition in Proposition~\ref{sufficient}:
\begin{equation*}
\begin{aligned}
&\frac{w(A'=a_k, Y'=1)}{w(A'=a_k, Y'=0)} = \frac{|A'=a_k, Y'=0|}{|A'=a_k, Y'=1|} = 1,& \forall a_k \in A'.
\end{aligned}
\end{equation*}
As a result, with the synthetic training data, it is possible for Fair-SMOTE~\cite{chakraborty2021bias} to achieve equalized odds and it is guaranteed to achieve $smAOD=0$ on its training data $(X', A', Y')$. However, by generating synthetic training data, Fair-SMOTE may include unrealistic training examples and also requires longer pre-processing time.

\subsection{Proposed Algorithms}\label{fairbalance}

\begin{figure}[!tb]
\includegraphics[width=\linewidth]{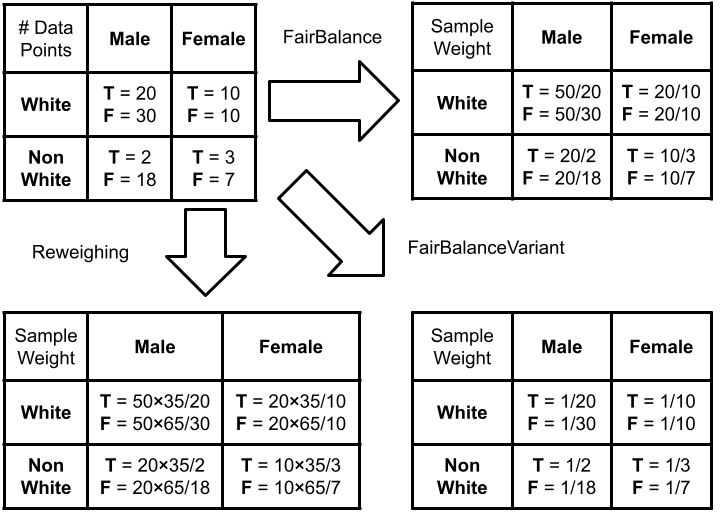}
\caption{ Demonstration of the preprocessing sample weights of FairBalance, FairBalanceVariant, and Reweighing. In this example, gender and race are the sensitive attributes, \textbf{T} and \textbf{F} are the two possible values of the dependent variable. The example training data consists of 50 white male, among which 20 have the label of \textbf{T} and 30 have the label of \textbf{F}. The FairBalance algorithm will assign a sample weight of 50/20 to each white male training data points with a \textbf{T} label and a sample weight of 50/30 to each white male training data points with a \textbf{F} label.  }
\label{fig:fairbalance}
\end{figure}

 Inspired by the reweighing algorithm~\cite{kamiran2012data} and based on the necessary condition in Proposition~\ref{necessary} and the sufficient condition in Proposition~\ref{sufficient},   we propose two pre-processing algorithms FairBalance and FairBalanceVariant with the following weighting mechanisms:
\begin{equation}\label{eq:fairbalance}
\begin{aligned}
&w_{FB}(A=a_k, Y=y_i) = \frac{|A=a_k|}{|A=a_k, Y=y_i|}\\
&w_{FBV}(A=a_k, Y=y_i) = \frac{1}{|A=a_k, Y=y_i|}\\
&\forall a_k \in A, \quad \forall y_i \in Y,
\end{aligned}
\end{equation}
Note that, both FairBalance and FairBalanceVariant satisfy the necessary condition in Proposition~\ref{necessary} and the sufficient condition in Proposition~\ref{sufficient}:
\begin{equation}\label{eq:satisfy}
\begin{aligned}
&\frac{w_{FB}(A=a_k, Y=1)}{w_{FB}(A=a_k, Y=0)} = \frac{w_{FBV}(A=a_k, Y=1)}{w_{FBV}(A=a_k, Y=0)}\\
= &\frac{|A=a_k, Y=0|}{|A=a_k, Y=1|} \quad \forall a_k \in A.
\end{aligned}
\end{equation}
It can be easily seen that the computational cost of the proposed algorithms are both $O(n)$ based on \eqref{eq:fairbalance}.  Figure~\ref{fig:fairbalance} demonstrates how the weights are calculated for FairBalance, FairBalanceVariant, and Reweighing. The differences between these three approaches are whether the original size difference in each demographic group $a_k\in A$ and the original class distribution are preserved. We will compare the performance of FairBalance, FairBalanceVariant, and Reweighing on real world test data to determine which approach is preferred.

\begin{table*}[!t]
\centering
\caption{Description of the data sets used in the experiment.}
\label{tab:dataset}
\setlength\tabcolsep{6pt}
\begin{tabular}{|p{0.18\textwidth}|c|c|cc|cc|}
\hline
\rowcolor[HTML]{C0C0C0} 
\textbf{Dataset}    & \textbf{\#Rows} & \textbf{\#Cols} & \multicolumn{2}{c|}{\cellcolor[HTML]{C0C0C0}\textbf{Protected Attribute}}                                                                                        & \multicolumn{2}{c|}{\cellcolor[HTML]{C0C0C0}\textbf{Class Label}}                      \\ \hline
\rowcolor[HTML]{C0C0C0} 
\textbf{}           & \textbf{}       & \textbf{}       & \multicolumn{1}{c|}{\cellcolor[HTML]{C0C0C0}\textbf{Privileged}}                       & \textbf{Unprivileged}                                                   & \multicolumn{1}{c|}{\cellcolor[HTML]{C0C0C0}\textbf{Favorable}} & \textbf{Unfavorable} \\ \hline
Adult Census Income~\cite{ADULT} & 48,842          & 14              & \multicolumn{1}{c|}{\begin{tabular}[c]{@{}c@{}}Sex-Male\\ Race-White\end{tabular}}     & \begin{tabular}[c]{@{}c@{}}Sex-Female\\ Race-Non-white\end{tabular}     & \multicolumn{1}{c|}{High Income}                                & Low Income           \\ \hline
Compas~\cite{COMPAS}             & 7,214           & 28              & \multicolumn{1}{c|}{\begin{tabular}[c]{@{}c@{}}Sex-Male\\ Race-Caucasian\end{tabular}} & \begin{tabular}[c]{@{}c@{}}Sex-Female\\ Race-Not Caucasian\end{tabular} & \multicolumn{1}{c|}{Did not reoffend}                           & Reoffended           \\ \hline
Heart Health~\cite{HEART}        & 297             & 14              & \multicolumn{1}{c|}{Age$< 60$}                                                         & Age$\ge 60$                                                                 & \multicolumn{1}{c|}{Not Disease}                                & Disease              \\ \hline
Bank Marketing~\cite{BANK}      & 45,211          & 16              & \multicolumn{1}{c|}{Age$>25$}                                                           & Age$\le 25$                                                               & \multicolumn{1}{c|}{Term Deposit - Yes}                         & Term Deposit - No    \\ \hline
German Credit Data~\cite{GERMAN}& 1,000          & 20              & \multicolumn{1}{c|}{\begin{tabular}[c]{@{}c@{}}Sex-Male\\ Age$>25$\end{tabular}}     & \begin{tabular}[c]{@{}c@{}}Sex-Female\\ Age$\le 25$\end{tabular}                                                               & \multicolumn{1}{c|}{Good Credit}                            & \multicolumn{1}{c|}{Bad Credit}    \\ \hline
Default of Credit Card Clients~\cite{DEFAULT}& 30,000          & 23              & \multicolumn{1}{c|}{\begin{tabular}[c]{@{}c@{}}Sex-Male\\ Age$>25$\end{tabular}}     & \begin{tabular}[c]{@{}c@{}}Sex-Female\\ Age$\le 25$\end{tabular}                                                               & \multicolumn{1}{c|}{No Default Payment}                            & \multicolumn{1}{c|}{Default Payment}    \\ \hline
Student Performance in Portuguese Language~\cite{STUDENT}& 395          & 32               & \multicolumn{1}{c|}{Sex-Male}     & \multicolumn{1}{c|}{Sex-Female}                                                            & \multicolumn{1}{c|}{Grade $\ge$ 10}                            & \multicolumn{1}{c|}{Grade $<$ 10}    \\ \hline
Student Performance in Mathematics~\cite{STUDENT}& 649          & 32               & \multicolumn{1}{c|}{Sex-Male}     & \multicolumn{1}{c|}{Sex-Female}                                                            & \multicolumn{1}{c|}{Grade $\ge$ 10}                            & \multicolumn{1}{c|}{Grade $<$ 10}    \\\hline
 SCUT-FBP5500~\cite{liang2018scut}    & 5,500          & 350$\times$350               & \multicolumn{1}{c|}{\begin{tabular}[c]{@{}c@{}}Sex-Male\\ Race-Asian\end{tabular}} & \begin{tabular}[c]{@{}c@{}}Sex-Female\\ Race-Caucasian\end{tabular}                                                            & \multicolumn{1}{c|}{Beauty Score $>$ 3}                            & \multicolumn{1}{c|}{Beauty Score $\le$ 3}    \\\hline
\end{tabular}
\end{table*}

\section{Experiments}
\label{sect:Experiments}

\subsection{Datasets}
\label{sect:data}

 For this study, we selected commonly used datasets in machine learning fairness to conduct our experiments. Starting with datasets seen in recent high-profile
papers~\cite{chakraborty2021bias,chakraborty2020fairway,lowy2021fermi,9286091}. This leads to the selection of the eight real world datasets (mostly from the UCI Machine Learning Repository~\cite{Dua:2019}) shown in the first eight rows of Table~\ref{tab:dataset}. All of these eight datasets were collected from real world data and represent a real problem. They also contain at least one sensitive attribute (four of the datasets contain two sensitive attributes) as independent variable. Experimenting on these dataset would generate a fair comparison between the proposed algorithm and the existing work. In addition to the eight tabular datasets, we also experimented on one real world face beauty image dataset SCUT-FBP5500~\cite{liang2018scut} (at Row 9 in Table~\ref{tab:dataset}) to demonstrate the generalizability of the proposed algorithms on complex deep neural networks in \textbf{RQ5}.  

\subsection{Evaluation}

The two machine learning fairness metrics mAOD and mEOD described in Section~\ref{sec:multiple} and their smoothed version smAOD and smEOD in Section~\ref{sec:smooth} are applied to evaluate the violation of equalized odds. In the meantime, accuracy is applied to evaluate the overall prediction performance:
\begin{equation}
\begin{aligned}
&\text{Accuracy} = P(\hat{Y}=Y).
\end{aligned}
\end{equation}
Since accuracy is largely affected by the classification threshold, we also apply the area under the ROC curve (AUC) shown in \eqref{eq:auc} to more comprehensively evaluate the utility of the learned model:
\begin{equation}\label{eq:auc}
\begin{aligned}
&\text{AUC} = \frac{\sum_{y_i=0}\sum_{y_j=1}\textbf{1}[f(x_i,a_i)<f(x_j,a_j)]}{|Y=1|\cdot|Y=0|}
\end{aligned}
\end{equation}
where $\textbf{1}[f(x_i,a_i)<f(x_j,a_j)]$ denotes an indicator function which returns 1 if $f(x_i,a_i)<f(x_j,a_j)$ otherwise returns 0. Runtime information of each treatment is also collected to reflect the computation overheads.

Each treatment is evaluated 30 times during experiments by each time randomly sampling 70\% of the data as training set and the rest as test set. Medians (50th percentile) and IQRs (75th percentile - 25th percentile) are collected for each performance metric since the resulting metrics do not follow a normal distribution. In addition, a nonparametric null-hypothesis significance testing (Mann–Whitney U test~\cite{mann1947test}) and a nonparametric effect size testing (Cliff's delta~\cite{cliff1993dominance}) are applied to check if one treatment performs significantly better than another in terms of a specific metric. A set of observations is considered to be significantly different from another set if and only if the null-hypothesis is rejected in the Mann–Whitney U test and the effect size in Cliff's delta is medium or large.
Similar to the Scott-Knott test~\cite{scott1974cluster}, rankings are also calculated to compare different treatments with nonparametric performance results. For each metric, the treatments are first sorted by their median values in that metric. Then, each pair of treatments is compared with the Mann–Whitney U test ($p\ge 0.05$) and Cliff's delta ($|\delta|<0.33$) to decide whether they belong to the same rank. 
Pseudo code of the ranking algorithm is shown in Algorithm~\ref{alg:ranking}.

\begin{algorithm}[!tbh]
\SetKwProg{Fn}{Function}{}{}
\SetKwInOut{Input}{Input}
\SetKwInOut{Output}{Output}
\SetKwInOut{Parameter}{Parameter}
\SetKwRepeat{Do}{do}{while}
\Input{\textbf{T}, performances to rank, a list of list.
}
\Output{\textbf{R}, rankings of the each treatment in \textbf{T}.}
\BlankLine
medians = []\\
\For{t $\in$ T}{
    medians.append(median(t))
}
asc = argsort(medians)\\
base = T[asc[0]]\\
rank = 0\\
R = []\\
R[asc[0]] = 0\\
\For{i=1, i<m, i++}{
    \If{MannWhitneyU\,(T[asc[i]],\,base) < 0.05 \& CliffsDelta\,(T[asc[i]],\,base) > 0.33}{
        rank = rank + 1\\
        base = T[asc[i]]
    }
    R[asc[i]] = rank
}
\Return R
\caption{Nonparametric ranking.}\label{alg:ranking}
\end{algorithm}

\subsection{Research Questions}\label{RQs}

Via experimenting on eight real world tabular datasets and one image processing dataset, we explore the following research questions:

\noindent \textbf{RQ1} Is the violation of equalized odds of the learned model positively related to the weighted class distribution differences across each demographic group in the training data--- is the ecessary condition in Proposition~\ref{necessary} valid?

\noindent \textbf{RQ2} Does balanced weighted class distribution in each demographic group lead to zero smAOD on the training data--- is the sufficient condition in Proposition~\ref{sufficient} valid?

\noindent \textbf{RQ3} Does the proposed algorithm outperform other \textbf{Category 3} algorithms in equalized odds?

\noindent \textbf{RQ4} Does removing certain training data (\textbf{Category 2} algorithms) help in achieving equalized odds?

To answer \textbf{RQ4}, the following two \textbf{Category 2} algorithms will be tested as well:

\textbf{Fairway}~\cite{chakraborty2020fairway}: First split the training data into partitions according to the values of one sensitive attribute, e.g. one partition with \textit{Sex=Male} and another partition with \textit{Sex=Female}. Then train a separate logistic regression model on each of the partitions. Next, the models are applied onto the training data and only the training data points which are predicted as the same class by all models are kept. Repeat this process if multiple sensitive attributes present.

\textbf{FairSituation}~\cite{chakraborty2021bias}: Fit a logistic regression model on the training data. Next, for each of the training data point $(x_i, a_i)$, create a counterpart of it $(x_i, A\neq a_i)$. Apply the model to predict on the training data point and its counterpart, the training data point is removed if the predictions are different.

\noindent \textbf{RQ5} Can the proposed algorithms be applied to solve more complicated problems such as image processing with deep neural networks--- is FairBalance model-agnostic or does it only apply to logistic regressors?

\subsection{Base Model Selection}\label{sect:model}

We utilized the first eight tabular datasets to explore RQ1 to RQ4. Same as the existing work utilizing the same tabular datasets~\cite{chakraborty2021bias,chakraborty2020fairway,9286091}, we fit a logistic regression model (implemented by scikit-learn with default hyper-parameters except for max\_iter=100,000) on each dataset with different pre-processing treatments including FairBalance. It has been shown that logistic regression models perform the best on these tabular datasets~\cite{chakraborty2021bias,chakraborty2020fairway,9286091}.

To explore RQ5 and demonstrate the generalizability of FairBalance, we employed a complex deep neural network called the VGG-16~\cite{simonyan2014very} model on an image processing dataset. Details of the VGG-16 model will be presented in Section~\ref{sect:RQ5}.

\section{Results}
\label{sec:results}

\begin{figure}[!tbh]
\includegraphics[width=\linewidth]{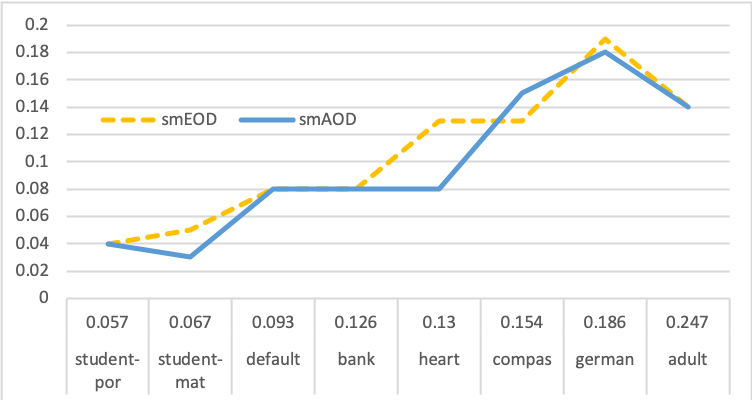}
\caption{Relationship between the violation of equalized odds of the learned model (measured as smEOD and smAOD at the y-axis) and the corresponding dataset's difference in class distributions of each demographics (measured as MaxDiff in \eqref{maxdiff} at the x-axis).}
\label{fig:data}
\end{figure}

\subsection{RQ1 Validate the necessary condition}

\textbf{RQ1} focuses on validating the necessary condition in Proposition~\ref{necessary} by empirically analyzing whether the violation of equalized odds of the learned model is positively related to the weighted class distribution differences across each demographic group in the training data. To this end, we plotted the smEOD and smAOD of a logistic regression classifier trained with uniform weights on eight datasets. In Figure~\ref{fig:data}, x-axis shows the MaxDiff values for each dataset calculated as \eqref{maxdiff} reflecting the maximum difference in class distributions across each demographic group.

\begin{equation}\label{maxdiff}
\begin{aligned}
&\text{MaxDiff} = \max_{a_i\in A}\frac{|A=a_i, Y=1|}{|A=a_i|}-\min_{a_j\in A}\frac{|A=a_j, Y=1|}{|A=a_j|}.
\end{aligned}
\end{equation}
As we can see from Figure~\ref{fig:data}, the extent of violation of equalized odds is positively related (not strictly since it is also related to the overall prediction accuracy of the learned model) to the MaxDiff values of the training data. This validates the necessary condition in Proposition~\ref{necessary} that the difference in class distributions across each demographic groups lead to the violation of equalized odds.

\begin{RQ}
{Answer to \textbf{RQ1}:} Yes. On eight real world datasets, we observe that the violation of equalized odds of the learned model is positively related to the weighted class distribution differences across each demographic group in the training data.
\end{RQ}

\begin{table}[!tb]
\caption{Model performance with FairBalance on the training data. Each cell shows the median (50th percentile) value, and the IQR value (75th percentile - 25th percentile) of the metric from 30 random repeats on a certain dataset.}
\label{tab:train}
\centering
\scriptsize
\setlength\tabcolsep{1.6pt}
\renewcommand{\arraystretch}{2}
\begin{tabular}{|p{0.075\textwidth} | P{0.06\textwidth} | P{0.06\textwidth} |
P{0.06\textwidth} | P{0.06\textwidth} | P{0.06\textwidth} | P{0.06\textwidth}|}
\hline
\rowcolor{black!10} Data        & Accuracy        & AUC             & mEOD            & mAOD            & smEOD           & smAOD           \\ \hline
Adult       &   0.82 (0.00) &   0.90 (0.00) &   0.06 (0.01) &   0.02 (0.01) &   0.08 (0.01) &   \cellcolor{black!10}0.01 (0.01) \\
Compas      &   0.69 (0.01) &   0.75 (0.00) &   0.08 (0.03) &   0.04 (0.02) &   0.03 (0.01) &   \cellcolor{black!10}0.02 (0.01) \\
Heart       &   0.85 (0.02) &   0.93 (0.02) &   0.03 (0.03) &   0.02 (0.02) &   0.03 (0.02) &   \cellcolor{black!10}0.00 (0.00) \\
Bank        &   0.85 (0.00) &   0.91 (0.00) &   0.07 (0.01) &   0.00 (0.00) &   0.04 (0.01) &   \cellcolor{black!10}0.00 (0.00) \\
German      &   0.75 (0.01) &   0.84 (0.01) &   0.13 (0.07) &   0.05 (0.04) &   0.09 (0.03) &   \cellcolor{black!10}0.02 (0.02) \\
Default     &   0.69 (0.01) &   0.72 (0.00) &   0.17 (0.02) &   0.12 (0.01) &   0.01 (0.00) &   \cellcolor{black!10}0.00 (0.00) \\
Student-por &   0.93 (0.02) &   0.98 (0.01) &   0.02 (0.02) &   0.02 (0.02) &   0.03 (0.01) &   \cellcolor{black!10}0.00 (0.00) \\
Student-mat &   0.97 (0.01) &   1.00 (0.00) &   0.03 (0.02) &   0.01 (0.01) &   0.03 (0.01) &   \cellcolor{black!10}0.00 (0.00)   \\\hline
\end{tabular}
\end{table}

\begin{figure*}[!tbh]
\centering
\begin{subfigure}{\textwidth}
  \centering
  \includegraphics[width=\linewidth]{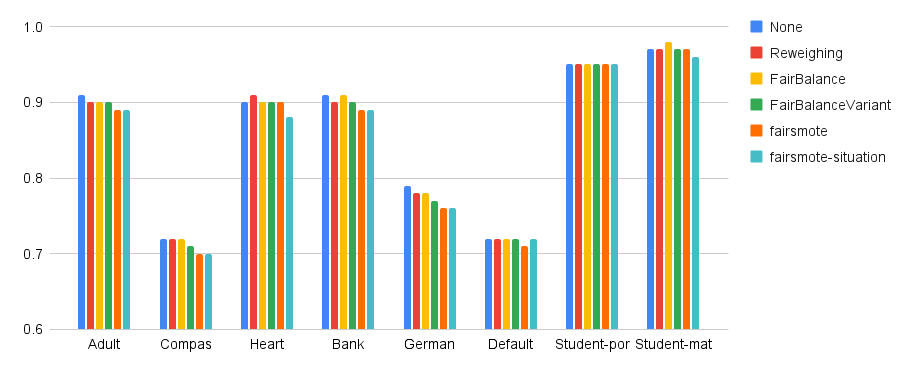}
  \caption{AUC of the ROC curve}
  \label{fig:auc1}
\end{subfigure}

\begin{subfigure}{\textwidth}
  \centering
  \includegraphics[width=\linewidth]{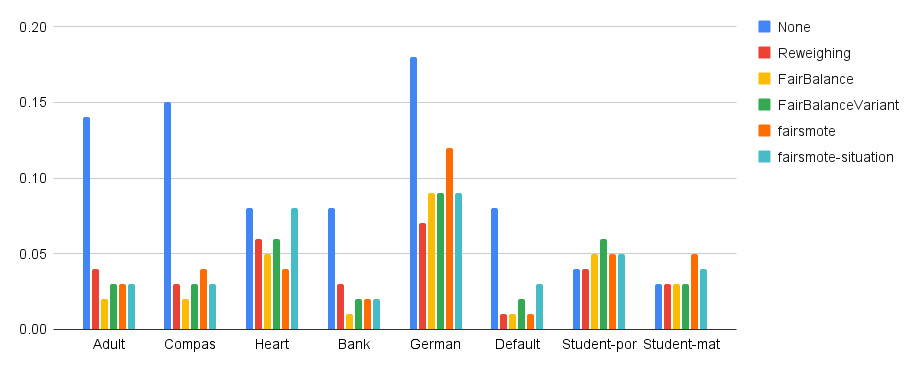}
  \caption{smAOD}
  \label{fig:smaod1}
\end{subfigure}
\caption{ Summarized results of median values for RQ3 (from Table~\ref{tab:adult1} to Table~\ref{tab:math1}).  }
\label{fig:test}
\end{figure*}

\subsection{RQ2 Validate the sufficient condition}

To validate the sufficient condition in Proposition~\ref{sufficient}, we apply FairBalance to multiply the weight $w_{FB}(a_k, y_i)$ in \eqref{eq:fairbalance} to each training data point. As shown in \eqref{eq:satisfy}, the weighted class distribution in each demographic group becomes balanced after applying the weights. Then we train a logistic regression model on the weighted training data and collect its training performance in Table~\ref{tab:train}. In consistency with the sufficient condition in Proposition~\ref{sufficient}, we can observe that the training smAODs are close to 0 on all eight datasets. The reason for smAODs on the Adult, Compas, and German datasets not strictly being 0 could be that, Assumption~\ref{Assumption 2} is not strictly valid for these datasets--- all three datasets have two sensitive attributes which can be correlated to each other. On the other hand, the smEODs are not always close to 0 even on the training data. 

\begin{RQ}
{Answer to \textbf{RQ2}:} Yes. Balanced weighted class distribution in each demographic group does lead to zero smAOD on the training data. However, it does not guarantee equalized odds on the training data since it does not always lead to zero smEOD. 
\end{RQ}

\begin{table*}[!tb]
\caption{Empirical results for RQ3 on the Adult Census Income dataset. Each cell shows (1) the ranking, (2) the median (50th percentile) value, and (3) the IQR value (75th percentile - 25th percentile) of the metric on a certain dataset. Colored cells are the ones with top rank r0.}
\label{tab:adult1}
\centering

\setlength\tabcolsep{3pt}
\renewcommand{\arraystretch}{2}
\begin{tabular}{|p{0.17\textwidth} | P{0.10\textwidth}P{0.10\textwidth} |
P{0.10\textwidth} P{0.10\textwidth} | P{0.10\textwidth} P{0.10\textwidth} | P{0.12\textwidth}|}
\hline
Treatment           & Accuracy        & AUC             & mEOD            & mAOD            & smEOD           & smAOD           & Runtime (secs)            \\ \hline
None                &  \cellcolor{black!10} r0: 0.85 (0.00) &  \cellcolor{black!10} r0: 0.91 (0.00) & r2: 0.19 (0.10) & r4: 0.14 (0.05) & r2: 0.14 (0.07) & r3: 0.14 (0.04) &  \cellcolor{black!10} r0: 1.18 (0.15)    \\
Reweighing          & r1: 0.84 (0.00) & r2: 0.90 (0.00) & r2: 0.17 (0.04) & r3: 0.08 (0.02) & r2: 0.12 (0.05) & r2: 0.04 (0.03) &  \cellcolor{black!10} r0: 1.23 (0.03)    \\
Fair-SMOTE           & r4: 0.81 (0.00) & r3: 0.89 (0.00) & r1: 0.07 (0.04) & r1: 0.05 (0.03) &  \cellcolor{black!10} r0: 0.08 (0.03) & r1: 0.03 (0.01) & r3: 142.83 (5.48)  \\
Fair-SMOTE-Situation & r3: 0.81 (0.01) & r2: 0.89 (0.00) &  \cellcolor{black!10} r0: 0.07 (0.04) & r2: 0.06 (0.01) &  \cellcolor{black!10} r0: 0.07 (0.03) & r1: 0.03 (0.02) & r3: 140.52 (14.06) \\
\textbf{FairBalance}         & r2: 0.81 (0.00) & r1: 0.90 (0.00) & r1: 0.08 (0.03) &  \cellcolor{black!10} r0: 0.03 (0.02) & r1: 0.09 (0.03) &  \cellcolor{black!10} r0: 0.02 (0.01) & r1: 1.23 (0.03)    \\
\textbf{FairBalanceVariant}  & r3: 0.81 (0.00) & r2: 0.90 (0.00) & r1: 0.08 (0.03) & r1: 0.05 (0.04) &  \cellcolor{black!10} r0: 0.07 (0.03) & r1: 0.03 (0.02) & r2: 1.34 (0.10)   \\\hline
\end{tabular}
\end{table*}

\begin{table*}[!tb]
\caption{Empirical results for RQ3 on the Compas dataset.}
\label{tab:compas1}
\centering
\setlength\tabcolsep{3pt}
\renewcommand{\arraystretch}{2}
\begin{tabular}{|p{0.17\textwidth} | P{0.10\textwidth}P{0.10\textwidth} |
P{0.10\textwidth} P{0.10\textwidth} | P{0.10\textwidth} P{0.10\textwidth} | P{0.12\textwidth}|}
\hline
Treatment           & Accuracy        & AUC             & mEOD            & mAOD            & smEOD           & smAOD           & Runtime (secs)         \\\hline
None                & \cellcolor{black!10} r0: 0.67 (0.01) & \cellcolor{black!10} r0: 0.72 (0.01) & r2: 0.23 (0.05) & r2: 0.31 (0.05) & r1: 0.13 (0.02) & r2: 0.15 (0.02) & \cellcolor{black!10} r0: 0.35 (0.03) \\
Reweighing          & r1: 0.67 (0.01) & \cellcolor{black!10} r0: 0.72 (0.02) & r1: 0.11 (0.05) & r1: 0.07 (0.05) & \cellcolor{black!10} r0: 0.03 (0.01) & \cellcolor{black!10} r0: 0.03 (0.02) & r1: 0.38 (0.04) \\
Fair-SMOTE           & r3: 0.65 (0.01) & r2: 0.70 (0.01) & \cellcolor{black!10} r0: 0.08 (0.04) & r1: 0.09 (0.05) & \cellcolor{black!10} r0: 0.03 (0.02) & r1: 0.03 (0.02) & r3: 7.86 (1.15) \\
Fair-SMOTE-Situation & r3: 0.65 (0.01) & r2: 0.70 (0.01) & \cellcolor{black!10} r0: 0.07 (0.04) & r1: 0.07 (0.04) & \cellcolor{black!10} r0: 0.03 (0.01) & r1: 0.03 (0.01) & r2: 6.31 (0.59) \\
\textbf{FairBalance}         & r2: 0.66 (0.00) & r1: 0.72 (0.01) & \cellcolor{black!10} r0: 0.07 (0.05) & \cellcolor{black!10} r0: 0.05 (0.04) & \cellcolor{black!10} r0: 0.03 (0.02) & \cellcolor{black!10} r0: 0.02 (0.02) & r1: 0.38 (0.05) \\
\textbf{FairBalanceVariant}  & r2: 0.66 (0.01) & r1: 0.71 (0.02) & r1: 0.10 (0.05) & r1: 0.07 (0.06) & \cellcolor{black!10} r0: 0.03 (0.01) & \cellcolor{black!10} r0: 0.03 (0.01) & r1: 0.38 (0.04)   \\\hline
\end{tabular}
\end{table*}

\begin{table*}[!tb]
\caption{Empirical results for RQ3 on the Heart Health dataset.}
\label{tab:heart1}
\centering
\setlength\tabcolsep{3pt}
\renewcommand{\arraystretch}{2}
\begin{tabular}{|p{0.17\textwidth} | P{0.10\textwidth}P{0.10\textwidth} |
P{0.10\textwidth} P{0.10\textwidth} | P{0.10\textwidth} P{0.10\textwidth} | P{0.12\textwidth}|}
\hline
Treatment           & Accuracy        & AUC             & mEOD            & mAOD            & smEOD           & smAOD           & Runtime (secs)         \\\hline
None                & \cellcolor{black!10} r0: 0.83 (0.05) & \cellcolor{black!10} r0: 0.90 (0.04) & r1: 0.11 (0.11) & \cellcolor{black!10} r0: 0.08 (0.08) & r1: 0.13 (0.11) & \cellcolor{black!10} r0: 0.08 (0.05) & \cellcolor{black!10} r0: 0.02 (0.00) \\
Reweighing          & \cellcolor{black!10} r0: 0.84 (0.05) & \cellcolor{black!10} r0: 0.91 (0.04) & \cellcolor{black!10} r0: 0.09 (0.13) & \cellcolor{black!10} r0: 0.08 (0.09) & \cellcolor{black!10} r0: 0.07 (0.08) & \cellcolor{black!10} r0: 0.06 (0.07) & r1: 0.02 (0.00) \\
Fair-SMOTE           & \cellcolor{black!10} r0: 0.81 (0.06) & \cellcolor{black!10} r0: 0.90 (0.06) & r1: 0.11 (0.12) & \cellcolor{black!10} r0: 0.05 (0.10) & \cellcolor{black!10} r0: 0.06 (0.11) & \cellcolor{black!10} r0: 0.06 (0.07) & r2: 0.14 (0.00) \\
Fair-SMOTE-Situation & r1: 0.79 (0.03) & r1: 0.88 (0.05) & \cellcolor{black!10} r0: 0.08 (0.10) & \cellcolor{black!10} r0: 0.08 (0.13) & \cellcolor{black!10} r0: 0.05 (0.08) & \cellcolor{black!10} r0: 0.06 (0.09) & r3: 0.14 (0.00) \\
\textbf{FairBalance}         & \cellcolor{black!10} r0: 0.83 (0.05) & \cellcolor{black!10} r0: 0.90 (0.04) & \cellcolor{black!10} r0: 0.09 (0.10) & \cellcolor{black!10} r0: 0.07 (0.09) & \cellcolor{black!10} r0: 0.07 (0.04) & \cellcolor{black!10} r0: 0.05 (0.06) & r1: 0.02 (0.00) \\
\textbf{FairBalanceVariant}  & \cellcolor{black!10} r0: 0.83 (0.05) & \cellcolor{black!10} r0: 0.90 (0.04) & \cellcolor{black!10} r0: 0.09 (0.10) & \cellcolor{black!10} r0: 0.06 (0.10) & \cellcolor{black!10} r0: 0.06 (0.09) & \cellcolor{black!10} r0: 0.06 (0.07) & \cellcolor{black!10} r0: 0.02 (0.00)   \\\hline
\end{tabular}
\end{table*}

\begin{table*}[!tb]
\caption{Empirical results for RQ3 on the Bank Marketing dataset.}
\label{tab:bank1}
\centering
\setlength\tabcolsep{3pt}
\renewcommand{\arraystretch}{2}
\begin{tabular}{|p{0.17\textwidth} | P{0.10\textwidth}P{0.10\textwidth} |
P{0.10\textwidth} P{0.10\textwidth} | P{0.10\textwidth} P{0.10\textwidth} | P{0.12\textwidth}|}
\hline
Treatment           & Accuracy        & AUC             & mEOD            & mAOD            & smEOD           & smAOD           & Runtime (secs)           \\ \hline
None                & \cellcolor{black!10} r0: 0.90 (0.00) & r1: 0.91 (0.00) & r3: 0.13 (0.08) & r2: 0.09 (0.04) & r1: 0.08 (0.03) & r2: 0.08 (0.02) & \cellcolor{black!10} r0: 0.98 (0.02)   \\
Reweighing          & \cellcolor{black!10} r0: 0.90 (0.00) & r1: 0.90 (0.01) & r2: 0.09 (0.06) & r1: 0.04 (0.03) & r1: 0.08 (0.02) & r1: 0.03 (0.02) & r3: 1.12 (0.02)   \\
Fair-SMOTE           & r3: 0.83 (0.01) & r3: 0.89 (0.01) & \cellcolor{black!10} r0: 0.05 (0.05) & \cellcolor{black!10} r0: 0.02 (0.03) & \cellcolor{black!10} r0: 0.03 (0.03) & \cellcolor{black!10} r0: 0.02 (0.02) & r4: 357.84 (3.24) \\
Fair-SMOTE-Situation & r2: 0.83 (0.01) & r3: 0.89 (0.01) & \cellcolor{black!10} r0: 0.06 (0.05) & \cellcolor{black!10} r0: 0.02 (0.02) & \cellcolor{black!10} r0: 0.03 (0.03) & \cellcolor{black!10} r0: 0.01 (0.02) & r4: 357.68 (1.94) \\
\textbf{FairBalance}         & r1: 0.84 (0.00) & \cellcolor{black!10} r0: 0.91 (0.00) & r1: 0.07 (0.05) & \cellcolor{black!10} r0: 0.02 (0.02) & \cellcolor{black!10} r0: 0.04 (0.03) & \cellcolor{black!10} r0: 0.01 (0.01) & r1: 1.05 (0.02)   \\
\textbf{FairBalanceVariant}  & r2: 0.83 (0.00) & r2: 0.90 (0.00) & \cellcolor{black!10} r0: 0.06 (0.05) & \cellcolor{black!10} r0: 0.02 (0.02) & \cellcolor{black!10} r0: 0.04 (0.04) & \cellcolor{black!10} r0: 0.02 (0.02) & r2: 1.10 (0.02)     \\\hline
\end{tabular}
\end{table*}

\begin{table*}[!tb]
\caption{Empirical results for RQ3 on the German Credit dataset.}
\label{tab:german1}
\centering
\setlength\tabcolsep{3pt}
\renewcommand{\arraystretch}{2}
\begin{tabular}{|p{0.17\textwidth} | P{0.10\textwidth}P{0.10\textwidth} |
P{0.10\textwidth} P{0.10\textwidth} | P{0.10\textwidth} P{0.10\textwidth} | P{0.12\textwidth}|}
\hline
Treatment           & Accuracy        & AUC             & mEOD            & mAOD            & smEOD           & smAOD           & Runtime (secs)         \\\hline
None                & \cellcolor{black!10} r0: 0.76 (0.03) & \cellcolor{black!10} r0: 0.79 (0.03) & r2: 0.25 (0.17) & r1: 0.29 (0.16) & r2: 0.19 (0.07) & r1: 0.18 (0.07) & \cellcolor{black!10} r0: 0.06 (0.00) \\
Reweighing          & \cellcolor{black!10} r0: 0.75 (0.03) & \cellcolor{black!10} r0: 0.78 (0.03) & \cellcolor{black!10} r0: 0.12 (0.08) & \cellcolor{black!10} r0: 0.16 (0.13) & \cellcolor{black!10} r0: 0.08 (0.06) & \cellcolor{black!10} r0: 0.07 (0.05) & r1: 0.07 (0.00) \\
Fair-SMOTE           & r1: 0.71 (0.04) & r1: 0.77 (0.04) & \cellcolor{black!10} r0: 0.15 (0.12) & \cellcolor{black!10} r0: 0.19 (0.11) & r1: 0.10 (0.07) & \cellcolor{black!10} r0: 0.11 (0.06) & r3: 1.85 (0.03) \\
Fair-SMOTE-Situation & r1: 0.71 (0.03) & r1: 0.77 (0.03) & r1: 0.17 (0.15) & \cellcolor{black!10} r0: 0.14 (0.10) & r1: 0.10 (0.09) & \cellcolor{black!10} r0: 0.08 (0.07) & r4: 1.91 (0.03) \\
\textbf{FairBalance}         & r1: 0.72 (0.04) & \cellcolor{black!10} r0: 0.78 (0.04) & r1: 0.21 (0.16) & \cellcolor{black!10} r0: 0.15 (0.11) & r1: 0.12 (0.10) & \cellcolor{black!10} r0: 0.09 (0.06) & r1: 0.07 (0.00) \\
\textbf{FairBalanceVariant}  & r2: 0.70 (0.03) & r1: 0.77 (0.03) & r1: 0.17 (0.16) & \cellcolor{black!10} r0: 0.18 (0.09) & r1: 0.11 (0.10) & \cellcolor{black!10} r0: 0.09 (0.05) & r2: 0.07 (0.00)    \\\hline
\end{tabular}
\end{table*}

\begin{table*}[!tb]
\caption{Empirical results for RQ3 on the Default of Credit Card Clients dataset.}
\label{tab:default1}
\centering
\setlength\tabcolsep{3pt}
\renewcommand{\arraystretch}{2}
\begin{tabular}{|p{0.17\textwidth} | P{0.10\textwidth}P{0.10\textwidth} |
P{0.10\textwidth} P{0.10\textwidth} | P{0.10\textwidth} P{0.10\textwidth} | P{0.12\textwidth}|}
\hline
Treatment           & Accuracy        & AUC             & mEOD            & mAOD            & smEOD           & smAOD           & Runtime (secs)           \\\hline
None                & \cellcolor{black!10} r0: 0.81 (0.00) & \cellcolor{black!10} r0: 0.72 (0.01) & r1: 0.04 (0.01) & r1: 0.06 (0.03) & r3: 0.08 (0.01) & r3: 0.08 (0.02) & \cellcolor{black!10} r0: 0.68 (0.01)   \\
Reweighing          & \cellcolor{black!10} r0: 0.81 (0.00) & r1: 0.72 (0.00) & \cellcolor{black!10} r0: 0.01 (0.01) & \cellcolor{black!10} r0: 0.05 (0.02) & \cellcolor{black!10} r0: 0.01 (0.01) & \cellcolor{black!10} r0: 0.01 (0.01) & r1: 0.86 (0.01)   \\
Fair-SMOTE           & r2: 0.68 (0.01) & r1: 0.72 (0.01) & r2: 0.16 (0.02) & r2: 0.12 (0.03) & r1: 0.01 (0.00) & \cellcolor{black!10} r0: 0.01 (0.01) & r3: 611.89 (4.19) \\
Fair-SMOTE-Situation & r3: 0.61 (0.02) & r1: 0.71 (0.01) & r2: 0.17 (0.07) & r2: 0.12 (0.05) & r2: 0.03 (0.01) & r2: 0.02 (0.01) & r3: 611.01 (2.59) \\
\textbf{FairBalance}         & r1: 0.69 (0.01) & r1: 0.72 (0.01) & r2: 0.17 (0.03) & r2: 0.13 (0.03) & r1: 0.02 (0.01) & \cellcolor{black!10} r0: 0.01 (0.01) & r1: 0.86 (0.02)   \\
\textbf{FairBalanceVariant}  & r2: 0.68 (0.01) & r1: 0.72 (0.01) & r2: 0.16 (0.03) & r2: 0.11 (0.03) & r1: 0.02 (0.01) & r1: 0.02 (0.01) & r2: 0.86 (0.01)      \\\hline
\end{tabular}
\end{table*}

\begin{table*}[!tb]
\caption{Empirical results for RQ3 on the Student Performance in Portuguese Language dataset.}
\label{tab:port1}
\centering
\setlength\tabcolsep{3pt}
\renewcommand{\arraystretch}{2}
\begin{tabular}{|p{0.17\textwidth} | P{0.10\textwidth}P{0.10\textwidth} |
P{0.10\textwidth} P{0.10\textwidth} | P{0.10\textwidth} P{0.10\textwidth} | P{0.12\textwidth}|}
\hline
Treatment           & Accuracy        & AUC             & mEOD            & mAOD            & smEOD           & smAOD           & Runtime (secs)         \\ \hline
None                & \cellcolor{black!10} r0: 0.92 (0.02) & \cellcolor{black!10} r0: 0.95 (0.02) & r1: 0.04 (0.02) & \cellcolor{black!10} r0: 0.05 (0.06) & r2: 0.04 (0.03) & \cellcolor{black!10} r0: 0.04 (0.03) & \cellcolor{black!10} r0: 0.05 (0.00) \\
Reweighing          & \cellcolor{black!10} r0: 0.91 (0.03) & \cellcolor{black!10} r0: 0.95 (0.02) & \cellcolor{black!10} r0: 0.02 (0.02) & \cellcolor{black!10} r0: 0.05 (0.09) & \cellcolor{black!10} r0: 0.01 (0.02) & \cellcolor{black!10} r0: 0.04 (0.05) & r2: 0.05 (0.01) \\
Fair-SMOTE           & r1: 0.88 (0.04) & \cellcolor{black!10} r0: 0.95 (0.02) & r1: 0.03 (0.03) & \cellcolor{black!10} r0: 0.05 (0.06) & r1: 0.03 (0.06) & \cellcolor{black!10} r0: 0.05 (0.07) & r3: 0.76 (0.01) \\
Fair-SMOTE-Situation & r1: 0.88 (0.03) & r1: 0.95 (0.02) & r1: 0.03 (0.04) & \cellcolor{black!10} r0: 0.06 (0.06) & r1: 0.03 (0.04) & \cellcolor{black!10} r0: 0.06 (0.06) & r4: 0.79 (0.01) \\
\textbf{FairBalance}         & r1: 0.89 (0.02) & r1: 0.95 (0.02) & r1: 0.04 (0.03) & \cellcolor{black!10} r0: 0.06 (0.06) & r1: 0.03 (0.04) & \cellcolor{black!10} r0: 0.05 (0.05) & r2: 0.05 (0.00) \\
\textbf{FairBalanceVariant}  & r1: 0.89 (0.02) & r1: 0.95 (0.01) & r1: 0.04 (0.05) & \cellcolor{black!10} r0: 0.06 (0.05) & r2: 0.04 (0.04) & \cellcolor{black!10} r0: 0.06 (0.08) & r1: 0.05 (0.00)     \\\hline
\end{tabular}
\end{table*}

\begin{table*}[!tb]
\caption{Empirical results for RQ3 on the Student Performance in Mathematics dataset.}
\label{tab:math1}
\centering
\setlength\tabcolsep{3pt}
\renewcommand{\arraystretch}{2}
\begin{tabular}{|p{0.17\textwidth} | P{0.10\textwidth}P{0.10\textwidth} |
P{0.10\textwidth} P{0.10\textwidth} | P{0.10\textwidth} P{0.10\textwidth} | P{0.12\textwidth}|}
\hline
Treatment           & Accuracy        & AUC             & mEOD            & mAOD            & smEOD           & smAOD           & Runtime (secs)         \\\hline
None                & \cellcolor{black!10} r0: 0.91 (0.04) & \cellcolor{black!10} r0: 0.97 (0.01) & \cellcolor{black!10} r0: 0.05 (0.05) & \cellcolor{black!10} r0: 0.04 (0.03) & \cellcolor{black!10} r0: 0.05 (0.03) & \cellcolor{black!10} r0: 0.03 (0.04) & \cellcolor{black!10} r0: 0.04 (0.00) \\
Reweighing          & \cellcolor{black!10} r0: 0.91 (0.03) & \cellcolor{black!10} r0: 0.97 (0.01) & \cellcolor{black!10} r0: 0.03 (0.04) & \cellcolor{black!10} r0: 0.03 (0.04) & \cellcolor{black!10} r0: 0.03 (0.05) & \cellcolor{black!10} r0: 0.03 (0.03) & r1: 0.04 (0.00) \\
Fair-SMOTE           & r1: 0.90 (0.04) & \cellcolor{black!10} r0: 0.97 (0.02) & r1: 0.05 (0.05) & \cellcolor{black!10} r0: 0.06 (0.07) & \cellcolor{black!10} r0: 0.06 (0.05) & \cellcolor{black!10} r0: 0.04 (0.05) & r2: 0.25 (0.00) \\
Fair-SMOTE-Situation & r1: 0.89 (0.04) & r1: 0.96 (0.02) & r1: 0.05 (0.05) & \cellcolor{black!10} r0: 0.05 (0.05) & \cellcolor{black!10} r0: 0.05 (0.06) & \cellcolor{black!10} r0: 0.03 (0.05) & r3: 0.27 (0.00) \\
\textbf{FairBalance}         & \cellcolor{black!10} r0: 0.91 (0.03) & \cellcolor{black!10} r0: 0.98 (0.01) & \cellcolor{black!10} r0: 0.03 (0.03) & \cellcolor{black!10} r0: 0.04 (0.04) & \cellcolor{black!10} r0: 0.04 (0.04) & \cellcolor{black!10} r0: 0.03 (0.04) & r1: 0.04 (0.00) \\
\textbf{FairBalanceVariant}  & \cellcolor{black!10} r0: 0.91 (0.03) & \cellcolor{black!10} r0: 0.97 (0.01) & r1: 0.05 (0.07) & \cellcolor{black!10} r0: 0.04 (0.06) & r1: 0.06 (0.06) & \cellcolor{black!10} r0: 0.03 (0.04) & r1: 0.04 (0.00)   \\\hline
\end{tabular}
\end{table*}

\begin{figure*}[!tbh]
\centering
\begin{subfigure}{\textwidth}
  \centering
  \includegraphics[width=\linewidth]{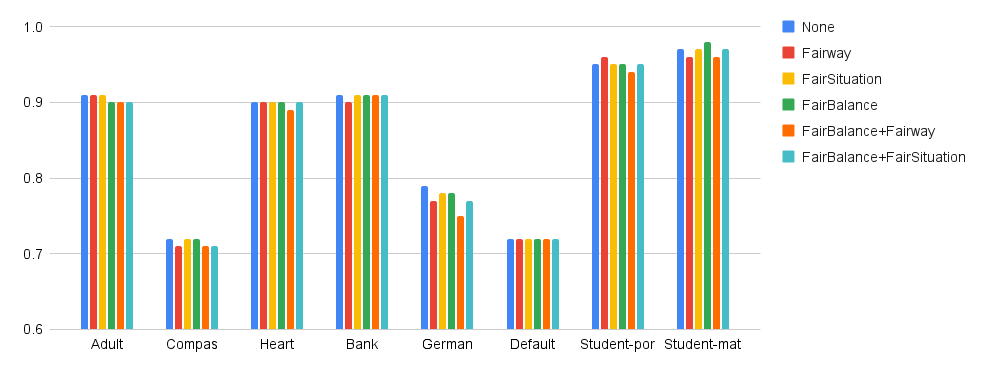}
  \caption{AUC of the ROC curve}
  \label{fig:auc1}
\end{subfigure}

\begin{subfigure}{\textwidth}
  \centering
  \includegraphics[width=\linewidth]{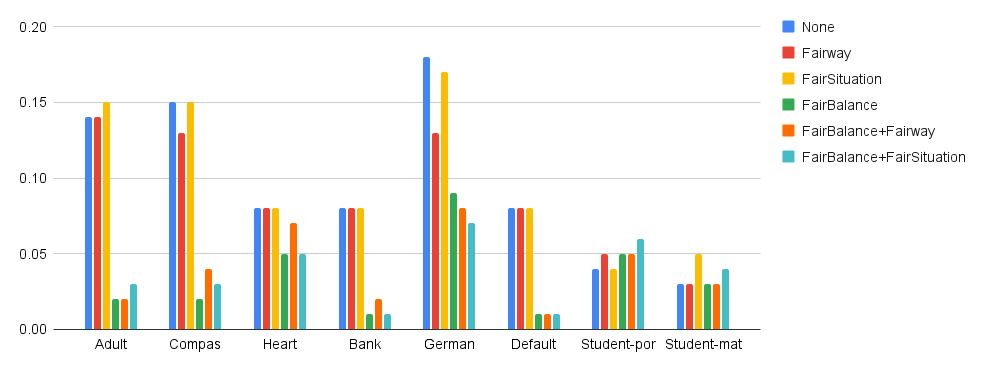}
  \caption{smAOD}
  \label{fig:smaod1}
\end{subfigure}
\caption{ Summarized results of median values for RQ4 (from Table~\ref{tab:adult2} to Table~\ref{tab:math2}).  }
\label{fig:test2}
\end{figure*}

\subsection{RQ3 Category 3 pre-processing}\label{sect:RQ3}

RQ3 tests the proposed pre-processing algorithms FairBalance and FairBalanceVariant against the state-of-the-art \textbf{Category 3} pre-processing algorithms Reweighing~\cite{kamiran2012data}, Fair-SMOTE~\cite{chakraborty2021bias}, and Fair-SMOTE-Situation~\cite{chakraborty2021bias}. Here, Fair-SMOTE-Situation is the combination of Fair-SMOTE and FairSituation. According to Chakraborty et al.~\cite{chakraborty2021bias} it first applies Fair-SMOTE to generate synthetic data points so that the training data is balanced as \eqref{eq:Fair-SMOTE}, then it applies FairSituation to remove data points failing the situation testing from the training data.  The SMOTE algorithm in this experiment is implemented with the same configuration as \cite{chakraborty2021bias}--- cr=0.8, f=0.8, and the number of neighbors is 3.   Each treatment is evaluated 30 times during the experiments by each time randomly sampling 70\% of the data as training set and the rest as test set.  Performances of each treatment on the test set are shown in Table~\ref{tab:adult1}-\ref{tab:math1} and are summarized in Figure~\ref{fig:test2}:  
\bi
\item
On most datasets (except for the Math dataset where the MaxDiff is close to 0 in Figure~\ref{fig:data}), equalized odds (measured by smEOD and smAOD) of the None treatment can be significantly improved after applying any of the pre-processing treatment. This aligns with the analysis that all these pre-processing treatments satisfy the necessary condition in Proposition~\ref{necessary}.
\item
FairBalance always achieves the best smAOD (ranked as r0) on every dataset. Following that, FairBalanceVariant and Fair-SMOTE are ranked r0 on 6 datasets and r1 on 2 datasets. Reweighing is ranked r0 on 6 datasets, r1 on the Bank Marketing dataset and r2 on the Adult Census Income dataset. Such results align with the analysis that FairBalance, FairBalanceVariant, and Fair-SMOTE satisfy the sufficient condition in Proposition~\ref{sufficient} which leads to better smAOD.
\item
In terms of smEOD, Reweighing (6 r0, 1 r1, and 1 r2) and Fair-SMOTE (5 r0 and 3 r1) outperform FairBalance (4 r0 and 4 r1) and the other algorithms.
\item
In terms of utility, FairBalance (4 r0 and 4 r1) and Reweighing (5 r0, 2 r1, and 1 r2) achieves the best AUC amongst the six treatments.
\item
In terms of runtime, FairBalance, FairBalanceVariant, and Reweighing have similar computational overheads (5-25\%), while Fair-SMOTE and Fair-SMOTE-Situation have much higher computational overheads.
\ei
Overall, the empirical results on the eight real world datasets are consistent with our analyses in Section~\ref{sec:Methodology}. Among the five tested pre-processing algorithms, we would recommend FairBalance since (1) it always achieves the best smAOD, (2) in terms of utility measured by AUC, it is also one of the best treatments, (3) it also has very small computational overhead. Note that, when smEOD and smAOD cannot be both satisfied, we value smAOD more since it is a more comprehensive metric (reflecting both the difference in true positive rate and false positive rate) than smEOD (which only relfects the difference in true positive rate). Note that, FairBalance achieves higher smAODs on the test sets than the training sets. This is due to the sampling bias which causes the training and test set not strictly following the same distribution. This is especially obvious on smaller datasets such as Heart, German, Student-Portuguese, and Student-Mathematics with $\le$1,000 samples.
 
\begin{RQ}
{Answer to \textbf{RQ3}:} Yes. The proposed algorithm FairBalance outperforms or on par with the other state-of-the-art \textbf{Category 3} pre-processing algorithms in terms of smAOD, AUC, and runtime. 
\end{RQ}

\begin{table*}[!tbh]
\caption{Empirical results for RQ4 on the Adult Census Income dataset.}
\label{tab:adult2}
\centering
\setlength\tabcolsep{3pt}
\renewcommand{\arraystretch}{2}
\begin{tabular}{|p{0.18\textwidth} | P{0.10\textwidth}P{0.10\textwidth} |
P{0.10\textwidth} P{0.10\textwidth} | P{0.10\textwidth} P{0.10\textwidth} | P{0.10\textwidth}|}
\hline
Treatment                 & Accuracy        & AUC             & mEOD            & mAOD            & smEOD           & smAOD           & Runtime (secs)         \\\hline
None & \cellcolor{black!10}r0: 0.85 (0.00) & \cellcolor{black!10}r0: 0.91 (0.00) & r2: 0.19 (0.10) & r3: 0.14 (0.05) & r2: 0.14 (0.07) & r2: 0.14 (0.04) & \cellcolor{black!10}r0: 1.27 (0.18)\\
Fairway                   &\cellcolor{black!10} r0: 0.85 (0.00) &\cellcolor{black!10} r0: 0.91 (0.00) & r2: 0.18 (0.07) & r3: 0.13 (0.03) & r2: 0.12 (0.05) & r2: 0.14 (0.03) & r3: 1.56 (0.05) \\
FairSituation             &\cellcolor{black!10} r0: 0.85 (0.00) &\cellcolor{black!10} r0: 0.91 (0.00) & r2: 0.19 (0.11) & r3: 0.14 (0.05) & r2: 0.16 (0.05) & r2: 0.15 (0.03) &\cellcolor{black!10} r0: 1.30 (0.02) \\
\textbf{FairBalance}               & r1: 0.81 (0.00) & r1: 0.90 (0.00) & r1: 0.08 (0.03) &\cellcolor{black!10} r0: 0.03 (0.02) & r1: 0.09 (0.03) &\cellcolor{black!10} r0: 0.02 (0.01) & r1: 1.39 (0.20) \\
FairBalance+Fairway       & r2: 0.80 (0.01) & r1: 0.90 (0.00) &\cellcolor{black!10} r0: 0.06 (0.03) & r1: 0.04 (0.02) &\cellcolor{black!10} r0: 0.08 (0.03) &\cellcolor{black!10} r0: 0.02 (0.01) & r4: 1.78 (0.04) \\
FairBalance+FairSituation & r2: 0.80 (0.01) & r1: 0.90 (0.00) &\cellcolor{black!10} r0: 0.07 (0.04) & r2: 0.06 (0.02) &\cellcolor{black!10} r0: 0.07 (0.02) & r1: 0.03 (0.02) & r2: 1.52 (0.03)  \\\hline
\end{tabular}
\end{table*}

\begin{table*}[!t]
\caption{Empirical results for RQ4 on the Compas dataset.}
\label{tab:compas2}
\centering
\setlength\tabcolsep{3pt}
\renewcommand{\arraystretch}{2}
\begin{tabular}{|p{0.18\textwidth} | P{0.10\textwidth}P{0.10\textwidth} |
P{0.10\textwidth} P{0.10\textwidth} | P{0.10\textwidth} P{0.10\textwidth} | P{0.10\textwidth}|}
\hline
Treatment                 & Accuracy        & AUC             & mEOD            & mAOD            & smEOD           & smAOD           & Runtime (secs)         \\\hline
None & \cellcolor{black!10} r0: 0.67 (0.01) & \cellcolor{black!10} r0: 0.72 (0.01) & r3: 0.23 (0.05) & r5: 0.31 (0.05) & r4: 0.13 (0.02) & r3: 0.15 (0.02) & \cellcolor{black!10}r0: 0.27 (0.01)\\
Fairway                   & r1: 0.66 (0.02) & r1: 0.71 (0.02) & r2: 0.18 (0.03) & r3: 0.25 (0.04) & r3: 0.10 (0.02) & r2: 0.13 (0.02) & r4: 0.41 (0.01) \\
FairSituation             &\cellcolor{black!10} r0: 0.67 (0.01) &\cellcolor{black!10} r0: 0.72 (0.01) & r3: 0.21 (0.03) & r4: 0.27 (0.05) & r4: 0.12 (0.03) & r3: 0.15 (0.04) & r2: 0.34 (0.01) \\
\textbf{FairBalance}               & r2: 0.66 (0.00) & r1: 0.72 (0.01) &\cellcolor{black!10} r0: 0.07 (0.05) &\cellcolor{black!10} r0: 0.05 (0.04) &\cellcolor{black!10} r0: 0.03 (0.02) &\cellcolor{black!10} r0: 0.02 (0.02) & r1: 0.31 (0.01) \\
FairBalance+Fairway       & r3: 0.65 (0.01) & r2: 0.71 (0.01) & r1: 0.12 (0.07) & r2: 0.10 (0.03) & r2: 0.06 (0.02) & r1: 0.04 (0.01) & r5: 0.43 (0.01) \\
FairBalance+FairSituation & r2: 0.66 (0.01) & r1: 0.71 (0.01) &\cellcolor{black!10} r0: 0.08 (0.06) & r1: 0.07 (0.03) & r1: 0.04 (0.02) &\cellcolor{black!10} r0: 0.03 (0.01) & r3: 0.38 (0.01)  \\\hline
\end{tabular}
\end{table*}

\begin{table*}[!t]
\caption{Empirical results for RQ4 on the Heart Health dataset.}
\label{tab:heart2}
\centering
\setlength\tabcolsep{3pt}
\renewcommand{\arraystretch}{2}
\begin{tabular}{|p{0.18\textwidth} | P{0.10\textwidth}P{0.10\textwidth} |
P{0.10\textwidth} P{0.10\textwidth} | P{0.10\textwidth} P{0.10\textwidth} | P{0.10\textwidth}|}
\hline
Treatment                 & Accuracy        & AUC             & mEOD            & mAOD            & smEOD           & smAOD           & Runtime (secs)         \\\hline
None & \cellcolor{black!10}r0: 0.83 (0.05) & \cellcolor{black!10}r0: 0.90 (0.04) & r1: 0.11 (0.11) & \cellcolor{black!10}r0: 0.08 (0.08) & r1: 0.13 (0.11) & \cellcolor{black!10}r0: 0.08 (0.05) & \cellcolor{black!10}r0: 0.02 (0.00)\\
Fairway                   &\cellcolor{black!10} r0: 0.82 (0.04) &\cellcolor{black!10} r0: 0.90 (0.03) & r1: 0.11 (0.17) & r1: 0.12 (0.15) &\cellcolor{black!10} r0: 0.12 (0.10) &\cellcolor{black!10} r0: 0.08 (0.11) & r3: 0.03 (0.00) \\
FairSituation             &\cellcolor{black!10} r0: 0.84 (0.03) &\cellcolor{black!10} r0: 0.90 (0.04) &\cellcolor{black!10} r0: 0.08 (0.10) &\cellcolor{black!10} r0: 0.09 (0.10) &\cellcolor{black!10} r0: 0.07 (0.08) &\cellcolor{black!10} r0: 0.08 (0.07) & r2: 0.02 (0.00) \\
\textbf{FairBalance}               &\cellcolor{black!10} r0: 0.83 (0.05) &\cellcolor{black!10} r0: 0.90 (0.04) &\cellcolor{black!10} r0: 0.09 (0.10) &\cellcolor{black!10} r0: 0.07 (0.09) &\cellcolor{black!10} r0: 0.07 (0.04) &\cellcolor{black!10} r0: 0.05 (0.06) & r1: 0.02 (0.00) \\
FairBalance+Fairway       &\cellcolor{black!10} r0: 0.82 (0.05) &\cellcolor{black!10} r0: 0.89 (0.03) &\cellcolor{black!10} r0: 0.06 (0.07) &\cellcolor{black!10} r0: 0.09 (0.06) &\cellcolor{black!10} r0: 0.07 (0.07) &\cellcolor{black!10} r0: 0.07 (0.09) & r3: 0.03 (0.00) \\
FairBalance+FairSituation &\cellcolor{black!10} r0: 0.82 (0.05) &\cellcolor{black!10} r0: 0.90 (0.03) &\cellcolor{black!10} r0: 0.08 (0.11) &\cellcolor{black!10} r0: 0.08 (0.07) &\cellcolor{black!10} r0: 0.06 (0.08) &\cellcolor{black!10} r0: 0.05 (0.06) & r2: 0.02 (0.00)  \\\hline
\end{tabular}
\end{table*}

\begin{table*}[!t]
\caption{Empirical results for RQ4 on the Bank Marketing dataset.}
\label{tab:bank2}
\centering
\setlength\tabcolsep{3pt}
\renewcommand{\arraystretch}{2}
\begin{tabular}{|p{0.18\textwidth} | P{0.10\textwidth}P{0.10\textwidth} |
P{0.10\textwidth} P{0.10\textwidth} | P{0.10\textwidth} P{0.10\textwidth} | P{0.10\textwidth}|}
\hline
Treatment                 & Accuracy        & AUC             & mEOD            & mAOD            & smEOD           & smAOD           & Runtime (secs)         \\\hline
None & \cellcolor{black!10} r0: 0.90 (0.00) & r1: 0.91 (0.00) & r1: 0.13 (0.08) & r1: 0.09 (0.04) & r1: 0.08 (0.03) & r1: 0.08 (0.02) & \cellcolor{black!10} r0: 0.97 (0.02)\\
Fairway                   & r1: 0.90 (0.00) & r2: 0.90 (0.01) & r1: 0.12 (0.06) & r1: 0.08 (0.03) & r1: 0.08 (0.04) & r1: 0.08 (0.02) & r2: 1.27 (0.02) \\
FairSituation             & r1: 0.90 (0.00) & r1: 0.91 (0.00) & r1: 0.12 (0.08) & r1: 0.08 (0.04) & r1: 0.08 (0.05) & r1: 0.08 (0.03) & r2: 1.27 (0.02) \\
\textbf{FairBalance}               & r2: 0.84 (0.00) &\cellcolor{black!10} r0: 0.91 (0.00) & r0: 0.07 (0.05) &\cellcolor{black!10} r0: 0.02 (0.02) &\cellcolor{black!10} r0: 0.04 (0.03) &\cellcolor{black!10} r0: 0.01 (0.01) & r1: 1.05 (0.02) \\
FairBalance+Fairway       & r4: 0.83 (0.00) &\cellcolor{black!10} r0: 0.91 (0.00) & r0: 0.08 (0.05) &\cellcolor{black!10} r0: 0.02 (0.03) &\cellcolor{black!10} r0: 0.05 (0.03) &\cellcolor{black!10} r0: 0.02 (0.02) & r3: 1.35 (0.05) \\
FairBalance+FairSituation & r3: 0.83 (0.00) &\cellcolor{black!10} r0: 0.91 (0.00) & r0: 0.06 (0.05) &\cellcolor{black!10} r0: 0.02 (0.02) &\cellcolor{black!10} r0: 0.05 (0.03) &\cellcolor{black!10} r0: 0.01 (0.01) & r4: 1.38 (0.03) \\\hline
\end{tabular}
\end{table*}

\begin{table*}[!t]
\caption{Empirical results for RQ4 on the German Credit dataset.}
\label{tab:german2}
\centering
\setlength\tabcolsep{3pt}
\renewcommand{\arraystretch}{2}
\begin{tabular}{|p{0.18\textwidth} | P{0.10\textwidth}P{0.10\textwidth} |
P{0.10\textwidth} P{0.10\textwidth} | P{0.10\textwidth} P{0.10\textwidth} | P{0.10\textwidth}|}
\hline
Treatment                 & Accuracy        & AUC             & mEOD            & mAOD            & smEOD           & smAOD           & Runtime (secs)         \\ \hline
None & \cellcolor{black!10} r0: 0.76 (0.03) & \cellcolor{black!10} r0: 0.79 (0.03) & r1: 0.25 (0.17) & r1: 0.29 (0.16) & r1: 0.19 (0.07) & r2: 0.18 (0.07) & \cellcolor{black!10} r0: 0.06 (0.00)\\
Fairway                   &\cellcolor{black!10} r0: 0.74 (0.03) & r1: 0.77 (0.05) &\cellcolor{black!10} r0: 0.21 (0.09) & r1: 0.23 (0.11) &\cellcolor{black!10} r0: 0.15 (0.07) & r1: 0.13 (0.09) & r4: 0.11 (0.00) \\
FairSituation             &\cellcolor{black!10} r0: 0.75 (0.02) &\cellcolor{black!10} r0: 0.78 (0.03) & r1: 0.26 (0.10) & r1: 0.28 (0.12) & r1: 0.18 (0.05) & r2: 0.17 (0.05) & r2: 0.08 (0.00) \\
\textbf{FairBalance}               & r1: 0.72 (0.04) &\cellcolor{black!10} r0: 0.78 (0.04) &\cellcolor{black!10} r0: 0.21 (0.16) &\cellcolor{black!10} r0: 0.15 (0.11) &\cellcolor{black!10} r0: 0.12 (0.10) &\cellcolor{black!10} r0: 0.09 (0.06) & r1: 0.07 (0.00) \\
FairBalance+Fairway       & r2: 0.68 (0.02) & r1: 0.75 (0.03) &\cellcolor{black!10} r0: 0.19 (0.11) &\cellcolor{black!10} r0: 0.15 (0.11) &\cellcolor{black!10} r0: 0.13 (0.07) &\cellcolor{black!10} r0: 0.08 (0.07) & r5: 0.11 (0.00) \\
FairBalance+FairSituation & r1: 0.70 (0.03) &\cellcolor{black!10} r0: 0.77 (0.04) &\cellcolor{black!10} r0: 0.23 (0.16) &\cellcolor{black!10} r0: 0.15 (0.11) &\cellcolor{black!10} r0: 0.15 (0.11) &\cellcolor{black!10} r0: 0.07 (0.09) & r3: 0.09 (0.00) \\\hline
\end{tabular}
\end{table*}

\begin{table*}[!t]
\caption{Empirical results for RQ4 on the Default of Credit Card Clients dataset.}
\label{tab:default2}
\centering
\setlength\tabcolsep{3pt}
\renewcommand{\arraystretch}{2}
\begin{tabular}{|p{0.18\textwidth} | P{0.10\textwidth}P{0.10\textwidth} |
P{0.10\textwidth} P{0.10\textwidth} | P{0.10\textwidth} P{0.10\textwidth} | P{0.10\textwidth}|}
\hline
Treatment                 & Accuracy        & AUC             & mEOD            & mAOD            & smEOD           & smAOD           & Runtime (secs)         \\\hline
None & \cellcolor{black!10}r0: 0.81 (0.00) & \cellcolor{black!10}r0: 0.72 (0.01) & \cellcolor{black!10} r0: 0.04 (0.01) & r1: 0.06 (0.03) & r1: 0.08 (0.01) & r1: 0.08 (0.02) & \cellcolor{black!10}r0: 0.68 (0.01)\\
Fairway                   & r2: 0.80 (0.00) &\cellcolor{black!10} r0: 0.72 (0.01) &\cellcolor{black!10} r0: 0.04 (0.01) &\cellcolor{black!10} r0: 0.06 (0.02) & r1: 0.08 (0.01) & r1: 0.08 (0.01) & r2: 0.86 (0.02) \\
FairSituation             &r1: 0.81 (0.00) &\cellcolor{black!10} r0: 0.72 (0.01) &\cellcolor{black!10} r0: 0.04 (0.02) & r1: 0.06 (0.02) & r1: 0.08 (0.01) & r1: 0.08 (0.02) & r1: 0.74 (0.01) \\
\textbf{FairBalance}               & r3: 0.69 (0.01) & r1: 0.72 (0.01) & r1: 0.17 (0.03) & r2: 0.13 (0.03) &\cellcolor{black!10} r0: 0.02 (0.01) &\cellcolor{black!10} r0: 0.01 (0.01) & r2: 0.85 (0.02) \\
FairBalance+Fairway       & r5: 0.66 (0.00) & r1: 0.72 (0.01) & r2: 0.18 (0.03) & r3: 0.14 (0.03) &\cellcolor{black!10} r0: 0.01 (0.01) &\cellcolor{black!10} r0: 0.01 (0.01) & r4: 1.05 (0.02) \\
FairBalance+FairSituation & r4: 0.67 (0.01) &\cellcolor{black!10} r0: 0.72 (0.01) & r2: 0.18 (0.02) & r3: 0.13 (0.04) &\cellcolor{black!10} r0: 0.01 (0.01) &\cellcolor{black!10} r0: 0.01 (0.01) & r3: 0.94 (0.01)\\\hline
\end{tabular}
\end{table*}

\begin{table*}[!t]
\caption{Empirical results for RQ4 on the Student Performance in Portuguese Language dataset.}
\label{tab:port2}
\centering
\setlength\tabcolsep{3pt}
\renewcommand{\arraystretch}{2}
\begin{tabular}{|p{0.18\textwidth} | P{0.10\textwidth}P{0.10\textwidth} |
P{0.10\textwidth} P{0.10\textwidth} | P{0.10\textwidth} P{0.10\textwidth} | P{0.10\textwidth}|}
\hline
Treatment                 & Accuracy        & AUC             & mEOD            & mAOD            & smEOD           & smAOD           & Runtime (secs)         \\\hline
None & \cellcolor{black!10} r0: 0.92 (0.02) & \cellcolor{black!10} r0: 0.95 (0.02) & \cellcolor{black!10} r0: 0.04 (0.02) & \cellcolor{black!10} r0: 0.05 (0.06) & \cellcolor{black!10} r0: 0.04 (0.03) & \cellcolor{black!10} r0: 0.04 (0.03) & \cellcolor{black!10} r0: 0.05 (0.00)\\
Fairway                   &\cellcolor{black!10} r0: 0.91 (0.02) &\cellcolor{black!10} r0: 0.96 (0.02) &\cellcolor{black!10} r0: 0.02 (0.02) &\cellcolor{black!10} r0: 0.06 (0.09) &\cellcolor{black!10} r0: 0.03 (0.02) &\cellcolor{black!10} r0: 0.05 (0.06) & r4: 0.07 (0.00) \\
FairSituation             &\cellcolor{black!10} r0: 0.92 (0.02) & r1: 0.95 (0.02) &\cellcolor{black!10} r0: 0.02 (0.03) &\cellcolor{black!10} r0: 0.06 (0.06) &\cellcolor{black!10} r0: 0.03 (0.04) &\cellcolor{black!10} r0: 0.04 (0.05) & r2: 0.07 (0.00) \\
\textbf{FairBalance}               & r1: 0.89 (0.02) & r1: 0.95 (0.02) & r1: 0.04 (0.03) &\cellcolor{black!10} r0: 0.06 (0.06) &\cellcolor{black!10} r0: 0.03 (0.04) &\cellcolor{black!10} r0: 0.05 (0.05) & r1: 0.05 (0.00) \\
FairBalance+Fairway       & r1: 0.89 (0.04) & r1: 0.94 (0.03) & r1: 0.04 (0.04) &\cellcolor{black!10} r0: 0.05 (0.05) &\cellcolor{black!10} r0: 0.04 (0.04) &\cellcolor{black!10} r0: 0.05 (0.04) & r5: 0.08 (0.00) \\
FairBalance+FairSituation & r1: 0.89 (0.02) & r1: 0.95 (0.02) & r1: 0.06 (0.07) &\cellcolor{black!10} r0: 0.09 (0.06) &\cellcolor{black!10} r0: 0.05 (0.05) &\cellcolor{black!10} r0: 0.06 (0.06) & r3: 0.07 (0.00)\\\hline
\end{tabular}
\end{table*}

\begin{table*}[!t]
\caption{Empirical results for RQ4 on the Student Performance in Mathematics dataset.}
\label{tab:math2}
\centering
\setlength\tabcolsep{3pt}
\renewcommand{\arraystretch}{2}
\begin{tabular}{|p{0.18\textwidth} | P{0.10\textwidth}P{0.10\textwidth} |
P{0.10\textwidth} P{0.10\textwidth} | P{0.10\textwidth} P{0.10\textwidth} | P{0.10\textwidth}|}
\hline
Treatment                 & Accuracy        & AUC             & mEOD            & mAOD            & smEOD           & smAOD           & Runtime (secs)         \\\hline
None & \cellcolor{black!10} r0: 0.91 (0.04) & \cellcolor{black!10} r0: 0.97 (0.01) & \cellcolor{black!10} r0: 0.05 (0.05) & \cellcolor{black!10} r0: 0.04 (0.03) & \cellcolor{black!10} r0: 0.05 (0.03) & \cellcolor{black!10} r0: 0.03 (0.04) & \cellcolor{black!10} r0: 0.04 (0.00)\\
Fairway                   & r1: 0.89 (0.04) & r1: 0.96 (0.02) &\cellcolor{black!10} r0: 0.05 (0.07) &\cellcolor{black!10} r0: 0.05 (0.05) & r1: 0.05 (0.04) &\cellcolor{black!10} r0: 0.03 (0.05) & r4: 0.06 (0.00) \\
FairSituation             &\cellcolor{black!10} r0: 0.91 (0.03) &\cellcolor{black!10} r0: 0.97 (0.02) & r1: 0.08 (0.07) & r1: 0.06 (0.04) & r1: 0.07 (0.06) &\cellcolor{black!10} r0: 0.05 (0.02) & r2: 0.06 (0.00) \\
\textbf{FairBalance}               &\cellcolor{black!10} r0: 0.91 (0.03) &\cellcolor{black!10} r0: 0.98 (0.01) &\cellcolor{black!10} r0: 0.03 (0.03) &\cellcolor{black!10} r0: 0.04 (0.04) &\cellcolor{black!10} r0: 0.04 (0.04) &\cellcolor{black!10} r0: 0.03 (0.04) & r1: 0.04 (0.00) \\
FairBalance+Fairway       & r1: 0.88 (0.04) & r1: 0.96 (0.02) & r1: 0.08 (0.08) &\cellcolor{black!10} r0: 0.06 (0.09) & r1: 0.06 (0.06) &\cellcolor{black!10} r0: 0.03 (0.04) & r5: 0.06 (0.00) \\
FairBalance+FairSituation & r1: 0.90 (0.03) &\cellcolor{black!10} r0: 0.97 (0.01) & r1: 0.05 (0.08) &\cellcolor{black!10} r0: 0.04 (0.03) &\cellcolor{black!10} r0: 0.05 (0.06) &\cellcolor{black!10} r0: 0.04 (0.04) & r3: 0.06 (0.00)\\\hline
\end{tabular}
\end{table*}

\subsection{RQ4 Category 2 pre-processing}

RQ3 has shown that FairBalance is the best \textbf{Category 3} pre-processing algorithm for smAOD and AUC. Inspired by the ensemble algorithms such as Chakraborty et al.~\cite{chakraborty2021bias}, RQ4 tests how the \textbf{Category 2} algorithms affect equalized odds and whether they can further improve the model's performance when applied in combination with FairBalance. 
 Performances of each treatment on the test set are shown in Table~\ref{tab:adult2}-\ref{tab:math2}
 and are summarized in Figure~\ref{fig:test2}:  
\bi
\item
Fairway and FairSituation cannot achieve comparable smAOD or smEOD with FairBalance on most of the datasets (5 out of 8). They do not have much improvement smAOD or smEOD over the None treatment either. Fairway only slightly improves smAOD and smEOD over the None treatment on 3 out of 8 datasets and FairSituation does so on only 1 dataset. This is consistent with our analysis since Fairway and FairSituation do not satisfy the necessary condition in Proposition~\ref{necessary}.
\item
Adding either Fairway of FairSituation to FairBalance only worthen the smAOD performance on at least one dataset. Both FairBalance+Fairway of FairBalance+FairSituation do not improve on other metrics such as AUC or smEOD comparing to just applying FairBalance.
\ei
Overall, the \textbf{Category 2} pre-processing algorithms do not improve equalized odds and there is little value in applying FairBalance along with them. 

\begin{RQ}
{Answer to \textbf{RQ4}:} No. Removing certain training data does not help in achieving better equalized odds. 
\end{RQ}

\subsection{RQ5 Model-agnostic}
\label{sect:RQ5}

To test the generalizability of the proposed algorithms on complex problems with deep neural networks, we experimented on the SCUT-FBP5500 dataset~\cite{liang2018scut}. This dataset consists of 5,500 face images from Male and Female, Caucassian and Asian. Sixty different raters manually rated each face image for their perceptive ratings ranging from 1 to 5 individually and the average ratings are used as the ground truth. In this experiment, ratings higher than 3 are considered as favorable and ratings lower than or equal to 3 are considered as unfavorable as shown in Table~\ref{tab:dataset}. The image dataset was split into 50\% for training, 20\% for validation, and 30\% for testing. A VGG-16~\cite{simonyan2014very} model with pre-trained weights on the ImageNet data is transferred to predict the average beauty ratings with the output layer being replaced as a dense layer of size 256 and a one node output layer. The model is optimized for binary cross-entropy loss with stochastic gradient descent in batches of size 150\footnote{The analysis in Section~\ref{sec:Methodology} applies and FairBalance will guarantee 0 smAOD on the training data when trained with full batches. However, due to memory limitation, we can only train with batch size of 150.}. The model is trained on 4 NVIDIA A100 Tensor Core GPUs with 320 Gigabytes memory and is repeated 10 times for each treatment. The results are shown in Table~\ref{tab:fbp}.

Empirical results from Table~\ref{tab:fbp} show that FairBalance is able to reduce both EOD and AOD significantly without damaging the accuracy and AUC for the image processing problem. This aligns with the results on the tabular datasets with logistic regression classifiers. 

\begin{RQ}
{Answer to \textbf{RQ5}:} Yes. FairBalance improved equalized odds on image processing problems with deep neural networks. This demonstrated that FairBalance is model-agnostic.
\end{RQ}

\begin{table*}[!tbh]
\caption{ Empirical results for RQ5 on the SCUT-FBP5500 dataset.  }
\label{tab:fbp}
\centering
\setlength\tabcolsep{3pt}
\renewcommand{\arraystretch}{2}
\begin{tabular}{|p{0.17\textwidth} | P{0.10\textwidth}P{0.10\textwidth} |
P{0.10\textwidth} P{0.10\textwidth} | P{0.10\textwidth} P{0.10\textwidth} | P{0.12\textwidth}|}
\hline
Treatment           & Accuracy        & AUC             & mEOD            & mAOD            & smEOD           & smAOD           & Runtime (secs)         \\\hline
None               & \cellcolor{black!10}0.89 (0.01) & \cellcolor{black!10}0.96 (0.00) & 0.14 (0.07) & 0.11 (0.03) & 0.14 (0.03) & 0.10 (0.02) & 686.61 (59.99)  \\
Reweighing         & \cellcolor{black!10}0.89 (0.01) & 0.95 (0.00) & 0.13 (0.06) & 0.08 (0.05) & 0.11 (0.05) & 0.06 (0.03) & 687.71 (0.20)   \\
\textbf{FairBalance}        & \cellcolor{black!10}0.89 (0.01) & \cellcolor{black!10}0.96 (0.00) & \cellcolor{black!10}0.09 (0.05) & \cellcolor{black!10}0.05 (0.02) & \cellcolor{black!10}0.07 (0.04) & \cellcolor{black!10}0.04 (0.01) & 687.68 (90.00)  \\
FairBalanceVariant & 0.88 (0.01) & 0.95 (0.01) & \cellcolor{black!10}0.09 (0.05) & 0.07 (0.05) & 0.08 (0.04) & 0.05 (0.03) & 627.69 (105.04)   \\\hline
\end{tabular}
\end{table*}

\section{Threats to validity}
\label{threats}

\textbf{Sampling Bias} - Conclusions may change if other datasets and classification models are used. Specifically, Zhang and Harman~\cite{zhang2021ignorance} showed that enlarging feature set of the data could improve both fairness and accuracy. We have attempted to reduce the sampling bias by using two different models logistic regression model and VGG-16 model as the base classifier and experimenting on nine different real world datasets (including one image processing dataset). 

\noindent\textbf{Evaluation Bias} - We focused on the equalized odds fairness notion and evaluated it with mEOD, mAOD, smEOD and smAOD. For scenarios where other types of fairness is required, e.g. demographic parity, the proposed algorithm does not apply. 

\noindent\textbf{Conclusion Validity} - Analyses in this work were made based on Assumption~\ref{Assumption 1} and \ref{Assumption 2}. Prior fairness studies made similar assumptions~\cite{chakraborty2021bias,Biswas_2020}. However, such assumptions may not always hold for data with human decisions. 

\noindent\textbf{External Validity} - This work focuses on classification problems  which are very common in AI software. We are currently working on extending it to regression problems.

\section{Conclusion and future work}
\label{sect:Conclusion}

 It is the responsibility of software developers to develop accountable and fair machine learning software that does not perform differently on different sensitive demographic groups (i.e. achieving equalized odds). This paper aims to help software developers design classifiers satisfying equalized odds by assigning different weights to the training data points. Through analysis of the training process of common classifiers, we first find,   \textit{equal weighted class distributions across each demographic group is a necessary condition for achieving equalized odds} (Proposition~\ref{necessary}). This is also validated empirically in RQ1 where we showed the extent of violation of equalized odds is positively related to the max difference in the class distributions, and that in RQ3 and RQ4, the pre-processing algorithms satisfying the necessary condition achieve better equalized odds than those do not. Our second finding is, \textit{when the weighted class distributions are balanced (1:1) in every demographic group, partial equalized odds (smAOD=0) can be guaranteed in the training data} (Proposition~\ref{sufficient}). This sufficient condition is empirically validated in RQ2 that with FairBalance balancing the training data, smAODs are close to 0 on every training datasets. Note that these two major findings are subject to Assumption~\ref{Assumption 1} and \ref{Assumption 2} in Section~\ref{sec:Methodology}. 

With the two findings, we proposed FairBalance, a pre-processing algorithm balancing the training data. With experiments on eight real world datasets, we show in RQ3 and RQ4 that FairBalance outperforms every other baseline in smAOD, and is on par with them in terms of utility (measured by AUC) and runtime.  Also demonstrated in RQ5 with a complex neural network model trained on an image processing dataset, FairBalance is model-agnostic itself and can be generalized to classifiers other than logistic regression. 

Overall, FairBalance is able to improve equalized odds of binary classifiers on training data by adjusting the sample weights. It can be applied with any unbiased predictor with zero mean of training errors and an intercept term $\theta^{(0)}$. The computational overhead of it is also linear. With all these advantages, we would recommend the application of FairBalance when developing machine learning software that is required to perform similarly across different sensitive demographic groups. This would solve problems such as the COMPAS example discussed in Section~\ref{sec:introduction}--- after balancing the class distributions with FairBalance, the false positive rates and true positive rates across white and black defendants become less different (measured as mEOD and mAOD in Table~\ref{tab:compas1}). Note that FairBalance cannot prevent the machine learning software from being unfair due to unfair training data labels. This would require data elicitation from domain experts.

Given the threats to validity discussed in Section~\ref{threats} and the above limitation, future work of this paper focuses on:
\bi
\item
How to detect and mitigate biased labels. Equalized odds is no longer reliable when ground truth labels can be biased. But this also creates an opportunity to isolate the bias inherited from the training data when applying FairBalance to mitigate the bias originated from the training process. When equalized odds is violated for a model trained with FairBalance, the violation is possibly due to the biased labels.
\item
How to mitigate potential ethical bias when the sensitive attributes are unknown or noisy. There are some existing work along this research~\cite{wang2020robust}, However, it remains as a major challenge for machine learning fairness.
\item
How to generalize this work to regression problems when the dependent variable and the sensitive attributes can be continuous. This requires a generalized definition of equalized odds for the regression problems.
\ei


%


\section*{Acknowledgment}
This work is partially funded by NSF grant 2245796.


\ifCLASSOPTIONcaptionsoff

  \newpage
\fi



\bibliographystyle{IEEEtran}
%
\bibliography{mybib}

%

\vspace{-10 mm}
\begin{IEEEbiography}[{\includegraphics[width=1in,clip,keepaspectratio]{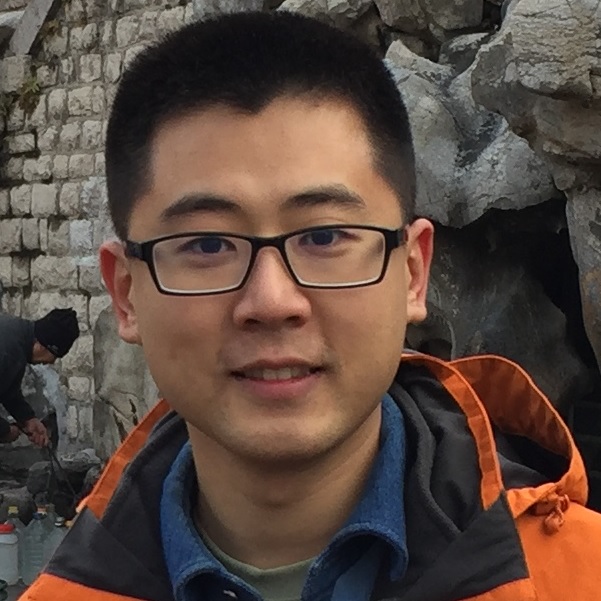}}]{Zhe Yu} (Ph.D. NC State University, 2020) is an
assistant professor in the Department of Software Engineering at Rochester Institute of Technology, where he teaches data mining
and software engineering. His research explores collaborations of human and machine learning algorithms that leads to
better performance and higher efficiency.
For more information, please visit \url{http://zhe-yu.github.io/}.
\end{IEEEbiography}
\vspace{-10 mm}
\begin{IEEEbiography}[{\includegraphics[width=1in,clip,keepaspectratio]{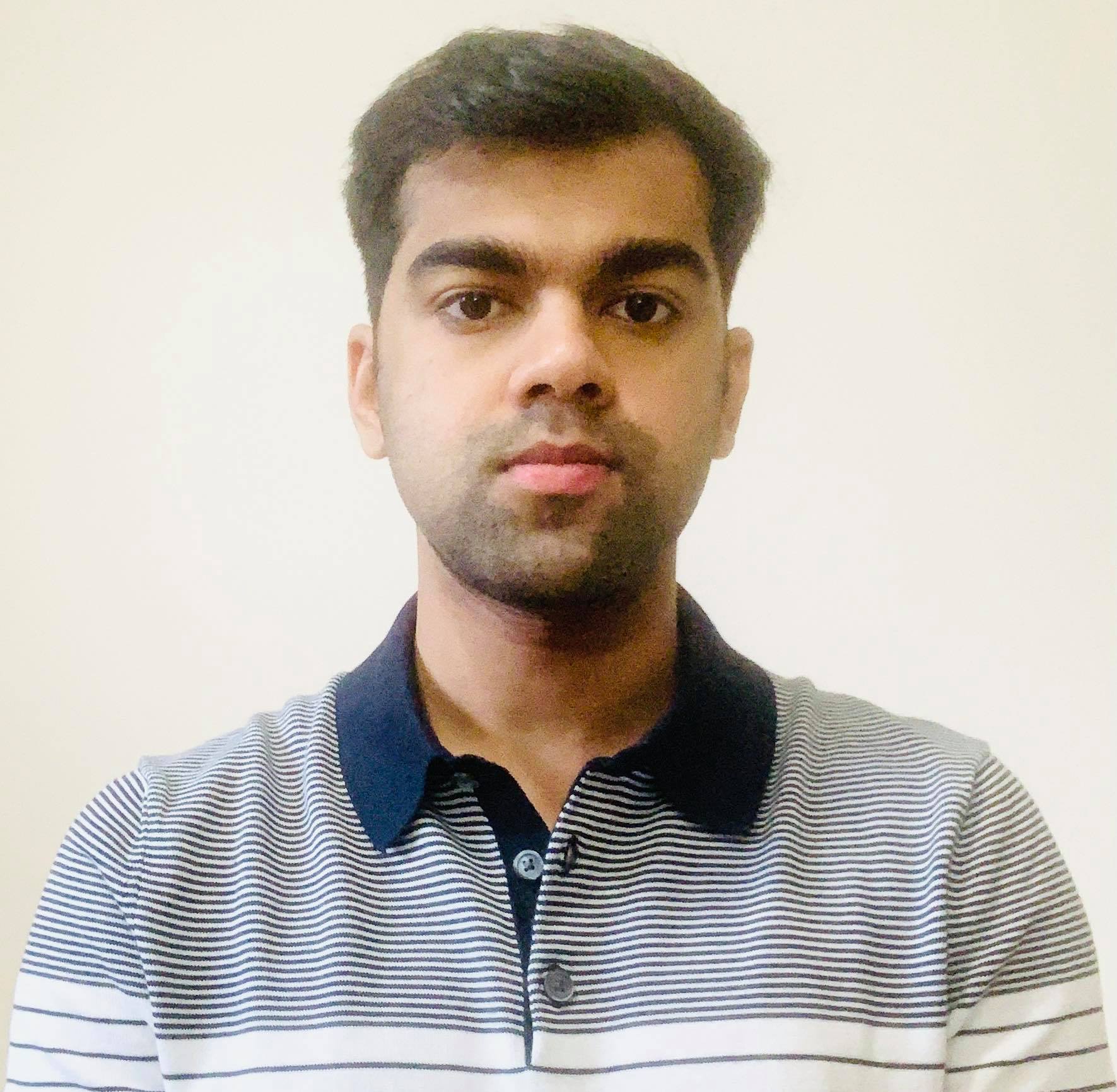}}]{Joymallya Chakraborty} is an Applied Scientist II at Amazon in Seattle. He completed his Ph.D. in Computer Science from North Carolina State University. His research interests include algorithmic bias, ML model optimization, interpretability \& explanation of black-box ML models. To know more about him, please visit https://joymallyac.github.io/
\end{IEEEbiography}
\vspace{-10 mm}
\begin{IEEEbiography}[{\includegraphics[width=1in,clip,keepaspectratio]{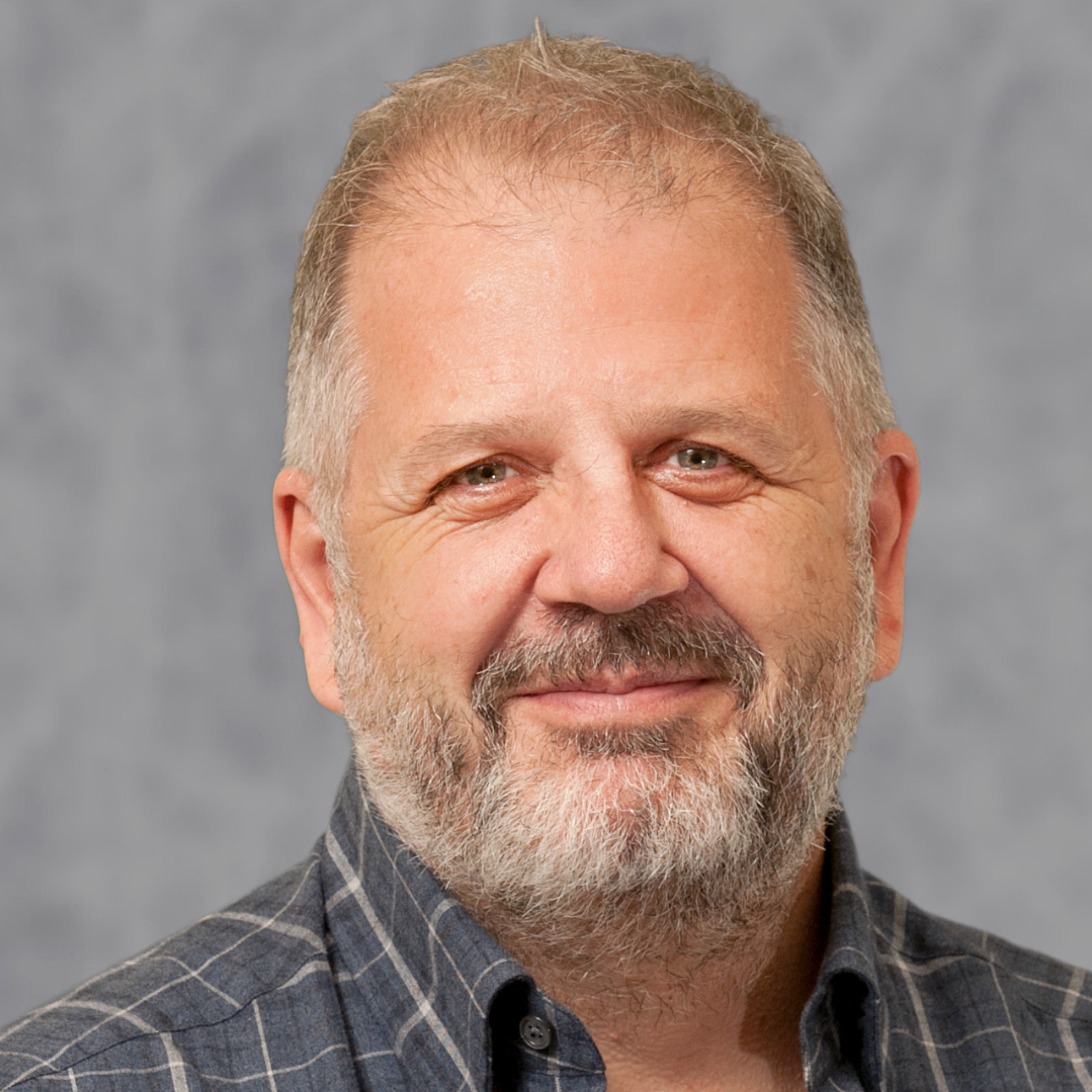}}]{Tim Menzies} (IEEE Fellow)
is a Professor in CS at NcState  His research interests include software engineering (SE), data mining, artificial intelligence, search-based SE, and open access science. \url{http://menzies.us}
\end{IEEEbiography}








\end{document}